\documentclass[sigconf]{acmart}

\AtBeginDocument{%
  }

\copyrightyear{2024}
\acmYear{2024}
\setcopyright{acmlicensed}\acmConference[KDD '24]{Proceedings of the 30th ACM SIGKDD Conference on Knowledge Discovery and Data Mining}{August 25--29, 2024}{Barcelona, Spain}
\acmBooktitle{Proceedings of the 30th ACM SIGKDD Conference on Knowledge Discovery and Data Mining (KDD '24), August 25--29, 2024, Barcelona, Spain}
\acmDOI{10.1145/3637528.3671836}
\acmISBN{979-8-4007-0490-1/24/08}

\usepackage{multirow}
\usepackage{xspace}
\usepackage{xcolor,colortbl}
\newcommand{\ours}{\texttt{EHRPD}\xspace}

\newcommand{\dmodel}{\textsf{P-DDPM}\xspace}
\newcommand{\unet}{\textsf{PU-Net}\xspace}

\definecolor{our}{rgb}{0.667, 0.863, 0.878}

\begin{document}

\title{Synthesizing Multimodal Electronic Health Records via Predictive Diffusion Models}

\author{Yuan Zhong}
\orcid{0009-0009-4427-5667}
\affiliation{%
  \institution{The Pennsylvania State University}
  \city{University Park}
  \state{PA}
  \postcode{16802}
  \country{USA}
}
\email{yfz5556@psu.edu}

\author{Xiaochen Wang}
\orcid{0009-0001-7699-3016}
\affiliation{%
  \institution{The Pennsylvania State University}
  \city{University Park}
  \state{PA}
  \postcode{16802}
  \country{USA}
}
\email{xcwang@psu.edu}

\author{Jiaqi Wang}
\orcid{0000-0002-9874-6622}
\affiliation{%
  \institution{The Pennsylvania State University}
  \city{University Park}
  \state{PA}
  \postcode{16802}
  \country{USA}
}
\email{jqwang@psu.edu}

\author{Xiaokun Zhang}
\orcid{0000-0002-9755-2471}
\affiliation{%
  \institution{Dalian University of Technology}
  \city{Dalian}
  \state{Liaoning}
  \postcode{116024}
  \country{China}
}
\email{dawnkun1993@gmail.com}

\author{Yaqing Wang}
\orcid{0000-0002-1548-0727}
\affiliation{%
  \institution{Purdue University}
  \city{West Lafayette}
  \state{IN}
  \postcode{47907}
  \country{USA}
}
\email{wang5075@purdue.edu}

\author{Mengdi Huai}
\orcid{0000-0001-6368-5973}
\affiliation{%
  \institution{Iowa State University}
  \city{Ames}
  \state{IA}
  \postcode{50011}
  \country{USA}
}
\email{mdhuai@iastate.edu}

\author{Cao Xiao}
\orcid{0000-0002-3869-6942}
\affiliation{%
  \institution{GE Healthcare}
  \city{Seattle}
  \state{WA}
  \postcode{}
  \country{USA}
}
\email{Cao.Xiao@gehealthcare.com}

\author{Fenglong Ma}
\orcid{0000-0002-4999-0303}
\affiliation{%
  \institution{The Penn State University}
  \city{University Park}
  \state{PA}
  \postcode{16802}
  \country{USA}
}
\email{fenglong@psu.edu}
\authornote{Corresponding authors.}

\renewcommand{\shortauthors}{Yuan Zhong et al.}

\begin{abstract}
Synthesizing electronic health records (EHR) data has become a preferred strategy to address data scarcity, improve data quality, and model fairness in healthcare. However, existing approaches for EHR data generation predominantly rely on state-of-the-art generative techniques like generative adversarial networks, variational autoencoders, and language models. These methods typically replicate input visits, resulting in inadequate modeling of temporal dependencies between visits and overlooking the generation of time information, a crucial element in EHR data. Moreover, their ability to learn visit representations is limited due to simple linear mapping functions, thus compromising generation quality.
To address these limitations, we propose a novel EHR data generation model called \ours. It is a diffusion-based model designed to predict the next visit based on the current one while also incorporating time interval estimation. To enhance generation quality and diversity, we introduce a novel time-aware visit embedding module and a pioneering predictive denoising diffusion probabilistic model (\dmodel). Additionally, we devise a predictive U-Net (\unet) to optimize \dmodel.
We conduct experiments on two public datasets and evaluate \ours from fidelity, privacy, and utility perspectives. The experimental results demonstrate the efficacy and utility of the proposed \ours in addressing the aforementioned limitations and advancing EHR data generation.
\end{abstract}

\ccsdesc[300]{Information systems~Data mining}
\ccsdesc[500]{Applied computing~Health informatics}
\ccsdesc[100]{Computing methodologies~Artificial intelligence}
\ccsdesc[300]{Computing methodologies~Neural networks}

\keywords{Electronic Health Records, Medical Data Synthesis, Diffusion Models, Multimodal Data Mining}


\maketitle

\section{Introduction}

\begin{figure}[t]
    \centering
\includegraphics[width=0.85\columnwidth]{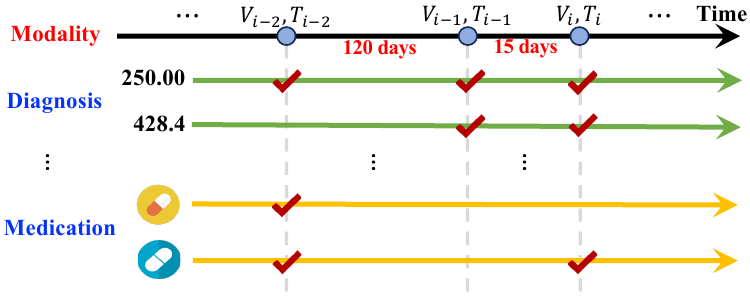}
    \caption{An example of multimodal EHR data, where $V_i$ denotes the visit information and $T_i$ represents its time.}
   \vspace{-0.1in}
    \label{fig:EHR_example}
\end{figure}

In healthcare, the utilization of Electronic Health Records (EHR) data is pivotal for advancing data-driven methodologies in both research and clinical practice~\cite{xiao2018opportunities}. EHR data possess distinctive characteristics, as illustrated in Figure~\ref{fig:EHR_example}, including sequential and temporal visit records, irregular time intervals between consecutive visits, and the presence of multiple modalities. However, effectively harnessing such intricate data encounters a significant challenge due to the scarcity of high-quality EHR datasets. To address this challenge, the generation of EHR data becomes an essential solution, providing a means to produce synthetic yet realistic supplements of patient data for constructing robust healthcare application models.

\begin{figure}[t]
    \centering
\includegraphics[width=0.9\columnwidth]{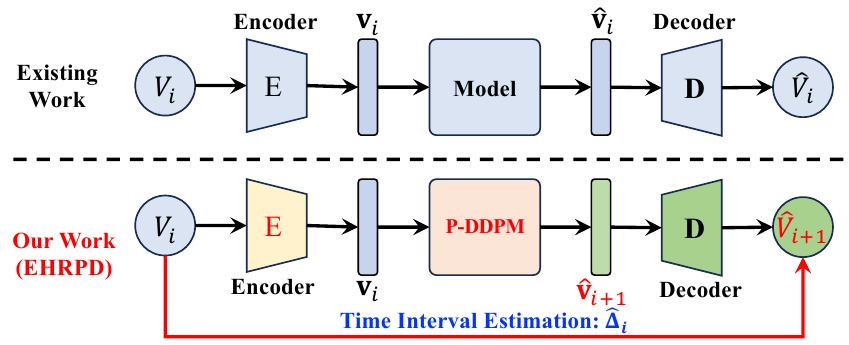}
    \vspace{-0.1in}
    \caption{Pipeline comparison between existing approaches and our proposed \ours.}
    \label{fig:pipeline}
    \vspace{-0.1in}
\end{figure}

Recent advancements in EHR data generation primarily depend on cutting-edge generative techniques, such as generative adversarial networks (GAN)~\cite{medGAN,ehrGAN,synteg}, variational autoencoders (VAE)~\cite{twin,eva}, and language models (LM)~\cite{promptehr,HALO}. These methodologies adhere to a common pipeline, as depicted in Figure~\ref{fig:pipeline}. This pipeline generally includes an encoder to encode visit $V_i$ into a representation $\mathbf{v}_i$, a generative model to generate the latent representation $\hat{\mathbf{v}}_i$, and a decoder to map $\hat{\mathbf{v}}_i$ to the generated visit $\hat{V}_i$. Despite their notable performance achievements, they still encounter several limitations.

$\bullet$ \textbf{Inadequate modeling of temporal dependencies between visits.}
Real EHR data, as shown in Figure~\ref{fig:EHR_example}, comprises visits ordered in time, with inherent temporal dependencies among them. However, existing methodologies employ a visit-replicating approach, generating a synthesis $\hat{V}_i$ for the input $V_i$ without explicitly addressing the temporal relationships between visits. An optimal generative model should inherently incorporate these temporal characteristics, such as directly using the current visit $V_i$ to generate the next visit $V_{i+1}$.

$\bullet$ \textbf{Failure to simultaneously generate time intervals between visits.} Current models aimed at generating detailed data frequently overlook a crucial aspect of patient healthcare: time information. As illustrated in Figure~\ref{fig:EHR_example}, time information is a vital component of EHR data, playing a significant role in modeling disease progression. Therefore, effectively capturing temporal dependencies between visits necessitates incorporating the modeling of time intervals between visits concurrently, enabling accurate characterization of patient health trajectories.

$\bullet$ \textbf{Limited capability in learning visit representations.} Owing to the discrete nature of certain EHR modalities like diagnosis codes, procedures, and medication codes, existing models embed the input visit $V_i$ into a continuous representation $\mathbf{v}_i$ first, as depicted in Figure~\ref{fig:pipeline}. The quality of the generated EHR data directly hinges on $\mathbf{v}_i$. However, current methods only utilize simple linear layers as the mapping function, potentially insufficient for representing the complexity of EHR data. Hence, there is a need to explore alternative approaches for learning visit representations.

$\bullet$ \textbf{Lack of a robust generation model to balance data diversity and quality.} From a modeling perspective, GAN-based approaches encounter the issue of model collapsing during training~\cite{medGAN,synteg,ehrGAN}, while VAE-based approaches rely on a strong Gaussian assumption that may not align well with EHR data~\cite{eva,twin}. LM-based approaches either depend on additional knowledge to generate diverse data, making it difficult to control data quality~\cite{promptehr}, or utilize autoregressive masked language training techniques to ensure quality but sacrifice diversity~\cite{HALO}. None of the existing models adequately address this challenging task. Therefore, the development of a powerful and comprehensive EHR generation model is urgently needed in healthcare.

\begin{figure*}[t]
    \centering
\includegraphics[width=0.85\textwidth]{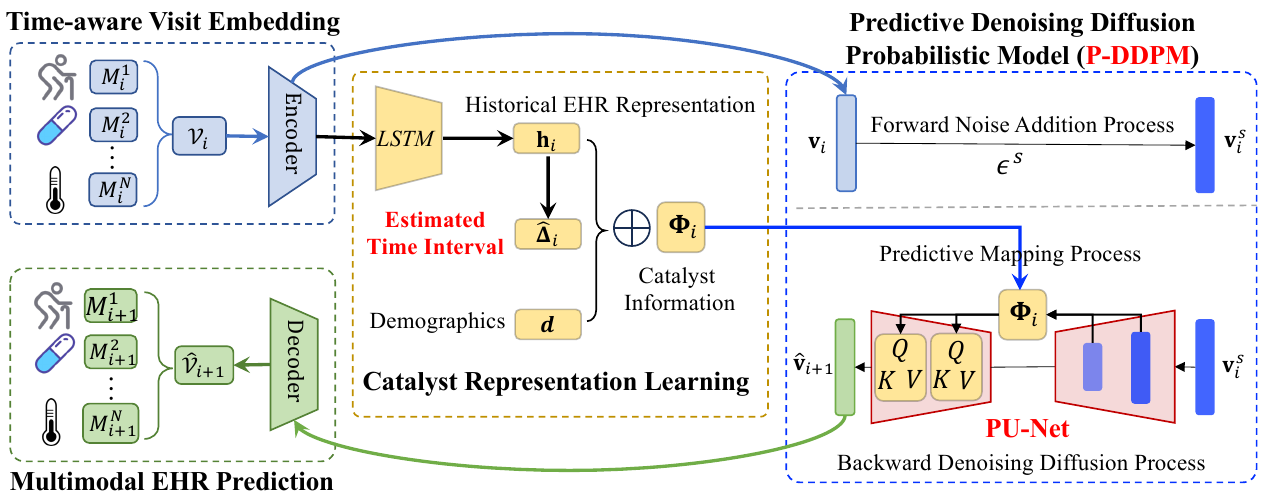}
    \caption{Overview of the proposed {\ours} model.}
    \label{fig:mainModel}
\end{figure*}

To comprehensively address the aforementioned limitations, we introduce \ours\footnote{\ours code repository: \url{https://anonymous.4open.science/r/EHRPD-465B}} in this paper, a diffusion-based model outlined in Figure~\ref{fig:mainModel}. Unlike existing approaches, \ours aims to capture the temporal characteristic of EHR data by generating the next visit $\hat{V}_{i+1}$ based on the current visit $V_i$.
Specifically, \ours takes multimodal EHR visit $V_i = \{M_i^1, \cdots, M_i^N\}$ as input, where $N$ denotes the number of modalities. Initially, \ours encodes the input $V_i$ using the designed \textbf{time-aware visit embedding} module, which facilitates the modeling of fine-grained code appearance patterns concerning time intervals when learning the visit embedding, denoted as $\mathbf{v}_i$.

The learned visit embedding $\mathbf{v}_i$ is then utilized to generate the latent representation of the subsequent visit $\hat{\mathbf{v}}_{i+1}$ via a novel \textbf{predictive denoising diffusion probabilistic model} (\dmodel). \dmodel comprises three key processes -- a forward noise addition process, a backward denoising diffusion process, and a predictive mapping process to encapsulate the temporality between visits in each diffusion step. To learn the latent representation $\hat{\mathbf{v}}_{i+1}$, we integrate the backward denoising diffusion process with a novel predictive U-Net (\unet).
The obtained representation $\hat{\mathbf{v}}_{i+1}$ is then employed to generate multimodal EHR data $\hat{V}_{i+1}$ through the decoding \textbf{EHR prediction} module.
The \textbf{catalyst representation learning} module is dedicated to estimating the time interval between $V_i$ and $\hat{V}_{i+1}$, as well as assembling the catalyst information $\boldsymbol{\Phi}_i$ used in \unet, including demographics $\mathcal{D}$, historical EHR representation $\mathbf{h}_i$, and the estimated time interval embedding $\hat{\boldsymbol{\Delta}}_i$.

In summary, the proposed \ours not only addresses the modeling of temporality between visits but also facilitates the simultaneous estimation of time intervals. Furthermore, the introduced time-aware visit embedding module can learn comprehensive visit embeddings by explicitly capturing code appearance patterns. 
Additionally, the design of \dmodel inherits the robustness of existing DDPMs~\cite{tabddpm,meddiffusion} while leveraging the diversity and quality of the generated EHR data through the noise addition and denoising process via the proposed \unet.
Finally, extensive experiments are conducted on MIMIC-III and Breast Cancer Trial datasets to validate the proposed \ours from fidelity, privacy, and utility perspectives. Experimental results demonstrate the superiority of \ours in EHR generation.

\section{Related Work}

\noindent\textbf{EHR Data Generation.}
To generate synthetic medical data to alleviate data scarcity, data generation methods in the medical domain that are equipped with GAN~\cite{medGAN,ehrGAN,synteg,DBLP:conf/kdd/ZhouXWNKAH21}, VAE~\cite{twin,eva}, LM~\cite{promptehr,HALO}, and DDPM~\cite{tabddpm,meddiffusion} have shown great success from their debut. Earlier methods~\cite{medGAN,ehrGAN} perform visit-level code aggregation to produce one or a few feature vectors and generate synthetic ones with GAN. However, this summarization would inevitably lose temporal dynamics and lead to inferior performance. To address this problem, recent work~\cite{eva,synteg,promptehr,twin,ehrGAN,HALO} aims to generate EHR data on the visit level. These fine-grained methods learn and leverage the hidden visit-wise relationship in EHR data with sequential learning techniques such as Tansformer, achieving state-of-the-art results. However, these methods ignore the time information of the patient's visit that contains crucial information such as disease progression and thus are suboptimal in their performance.



\smallskip
\noindent\textbf{Denoising Diffusion Probabilistic Models.}
The diffusion model has achieved considerable success in various tasks. One of its primary applications is image generation, as demonstrated in works~\cite{li2022srdiff,rombach2022high,saharia2022image}. It can also be adapted to time series forecasting and imputation~\cite{rasul2021autoregressive,tashiro2021csdi,rombach2022high}. Besides, the Discrete Diffusion Model~\cite{austin2021structured} adapts diffusion to discrete data space and fosters work such as~\cite{li2022diffusion,gong2022diffuseq}. Specific to the medical domain, the diffusion model has been used to generate healthcare data, including works such as~\cite{tabddpm,meddiffusion,he2023meddiff,naseer2023scoehr}. However, these methods are either task-oriented or not designed for sequential EHR generation.

\section{Methodology}

This work is dedicated to the generation of realistic high-dimensional, longitudinal, and multimodal Electronic Health Record (EHR) data. Given the sequential temporality inherent in EHR data, our objective is to simultaneously generate the next visit $\hat{V}_{i+1}$ and its associated time interval $\hat{\Delta}_{i}$ (i.e., $\hat{\Delta}_{i}=\hat{T}_{i+1} -T_i$), where $\hat{T}_{i+1}$ means the estimated time for $\hat{V}_{i+1}$. This generation process relies on the entire historical visit sequence $\mathcal{V}_{1:i}=[(V_1, T_1), \cdots, (V_i, T_i)]$ in conjunction with demographic information $\mathcal{D}$. Mathematically, this is represented as:
\begin{equation}\label{eq:overview}
    \{\hat{V}_{i+1}, \hat{\Delta}_i\} = g(\mathcal{V}_{1:i}, \mathcal{D}),
\end{equation}
where $g(\cdot)$ denotes the generation function. Each visit $V_i = \{M_i^1, \cdots, M_i^N\}$ encompasses $N$ modalities, including diagnosis codes, medication codes, lab test items, and so on.

To achieve this, we propose a novel diffusion-based EHR generation model, referred to as \ours, illustrated in Figure~\ref{fig:mainModel}. This model comprises four main modules: (1) time-aware visit embedding, (2) predictive denoising diffusion probabilistic model, (3) catalyst representation learning, and (4) multimodal EHR prediction. In the following section, we provide a detailed explanation of each module.

\subsection{Time-aware Visit Embedding}\label{sec:code-visit-embedding}
The easiest way to embed each visit $V_i = \{M_i^1, \cdots, M_i^N\}$ is applying a linear mapping function for each modality $M_i^n$ on its modality-level binary representation $\mathbf{y}_{i}^{n} \in \{0,1\}^{|\mathcal{C}_n|}$. However, this approach ignores nuances of the code's evolution against time. To address this deficiency, we propose a time-aware visit embedding approach to capture the temporal evolution of medical code individually.


\noindent\underline{\textbf{Time-aware Code Embedding.}}
For the $j$-th code $c_{i}^{n,j}$ that appears in the $n$-th modality, we record its most recent appearance time, which is then subtracted by $T_i$ to obtain the code-level time gap denoted as $\tau_i^{n,j}$. $\tau_i^{n,j}=0$ for the first visit. A smaller $\tau_i^{n,j}$ usually indicates a higher importance level for time-aware code embedding learning, which is described as follows:
\begin{equation}
    \mathbf{c}_i^{n,j} = \pi_i^{n,j}\text{MLP}_c([\mathbf{e}_i^{n,j};  \boldsymbol{\tau}_i^{n,j}]) + (1- \pi_i^{n,j})\mathbf{e}_i^{n,j},
\end{equation}
where $[;]$ denotes the concatenation operation, $\mathbf{e}_i^{n,j}$ is the basic code embedding, and $\boldsymbol{\tau}_i^{n,j}$ is the time gap embedding calculated by the positional embedding used in Transformer. $\pi_i^{n,j}$ is a gating indicator to decide whether to incorporate time gap information into code embedding learning, which is obtained via a Gumbel-Softmax layer as follows:
\begin{equation}
        \pi_i^{n,j} = \text{Binarize}\left( \frac{\exp\left({(\log(\mathbf{p}_i^{n,j}[0]) + G_0)}/{\eta}\right)}{\sum_{y=0}^{1} \exp\left({(\log(\mathbf{p}_i^{n,j}[y]) + G_y)}/{\eta}\right)} \right),
\label{eq:gumbelSofmax}
\end{equation}
where $\mathbf{p}_i^{n,j}$ is the softmax layer output on top of a linear function on the concatenated $[ \mathbf{e}_i^{n,j};  \boldsymbol{\tau}_i^{n,j}]$, $G$ is the noise following the Gumbel distribution, and $\eta$ is a hyperparameter.

\noindent\underline{\textbf{Visit Embedding.}}
We use a modality-level attention mechanism to learn the aggregated time-aware visit embedding as follows:
\begin{equation}\label{eq:visit_embedding}
    \begin{gathered}
        \mathbf{v}_i = \sum_{n=1}^N \psi_i^n \mathbf{z}_i^n,\\
        \boldsymbol{\psi}_i = \text{Softmax}\left(\text{MLP}_{\psi}([\mathbf{z}_i^1; \cdots; \mathbf{z}_i^N])\right),\\
        \mathbf{z}_i^n = \text{ReLU}\left(\text{MLP}_z \left(\sum_{j=1}^{|\mathcal{C}_n|}\mathbf{c}_i^{n,j}\right)\right).
    \end{gathered}
\end{equation}

\subsection{Predictive Denoising Diffusion Probabilistic Models (\dmodel)}\label{sec:pddpm}
Existing diffusion-based models, like DDPM~\cite{ddpm} and Glide~\cite{glide}, achieve generation by reconstructing the original input data. However, the task of EHR generation differs significantly from other tasks, as it aims to generate sequential, time-ordered EHR data using Eq.~\eqref{eq:overview} instead of reconstructing input data. 
To address this distinction, we propose a novel approach called the Predictive Denoising Diffusion Probabilistic Model (\dmodel) to generate the visit information $V_{i+1}$. Specifically, we treat the visit $V_i$ as a reactant and $V_{i+1}$ as the product. In generating $V_{i+1}$ using $V_i$, the aggregated information from $\mathcal{V}_{1:i}$ and $\mathcal{D}$ can be treated as the catalyst in \dmodel. Thus, the proposed \dmodel contains three components -- a forward noise addition process, a predictive mapping process, and a backward denoising diffusion process -- to tackle the challenges associated with sequential EHR data generation effectively.

\smallskip
\noindent\underline{\textbf{Forward Noise Addition Process.}}
The forward noise addition process is fixed to a Markov chain that gradually adds Gaussian noise to the representation of $V_i$ (i.e., $\mathbf{v}_i$ or  $\mathbf{v}_i^0$  detailed in Section~\ref{sec:code-visit-embedding} Eq.~\eqref{eq:visit_embedding}) as follows:
\begin{equation}
    \begin{gathered}
    q(\mathbf{v}_i^{1:S}|\mathbf{v}_i^0) = \prod_{s=1}^S q(\mathbf{v}_i^s|\mathbf{v}_i^{s-1}),\\
    q(\mathbf{v}_i^s|\mathbf{v}_i^{s-1}) = \mathcal{N}(\mathbf{v}_i^s;\sqrt{1-\beta_s} \mathbf{v}_i^{s-1},\beta_s\mathbf{I}),
\end{gathered}
\label{eq:forward}
\end{equation}
where $S$ is the number of diffusion steps, $q(\mathbf{v}_i^{1:S}|\mathbf{v}_i^0)$ is the approximate posterior, and $\beta_s$ is the variance schedule at step $s$.
Let $\alpha_s = 1-\beta_s$ and $\bar{\alpha}_s = \prod_{j=1}^s\alpha_j$, we can reparametrize the above Gausssian steps in Eq.~\eqref{eq:forward} to obtain the closed-form solution of $\mathbf{v}_i^s$ at any step $s$ without adding noise step by step as follows:
\begin{equation}
    \mathbf{v}_i^s = \sqrt{\alpha_s} \mathbf{v}_i^{s-1} + \sqrt{1-\alpha_s}\epsilon_{s-1} = \sqrt{\bar{\alpha}_s} \mathbf{v}_i^0 + \sqrt{1-\bar{\alpha}_s}\epsilon,
\label{eq:forwardRepa}
\end{equation}
where $\epsilon \in \mathcal{N}(\mathbf{0}, \mathbf{I})$. The details of the forward process can be found in Appendix Section~\ref{apd:forward}.

\smallskip
\noindent\underline{\textbf{Predictive Mapping Process.}} 
To generate the next visit $V_{i+1}$ using $V_i$, we need to first model the relationship between these two consecutive visits. In healthcare, such a relationship is usually modeled by a disease progress function, which is equivalent to a mapping function to predict $V_{i+1}$ using $V_i$ along with other information. Mathematically, we define such a predictive mapping function $f(\cdot)$ at each diffusion step as follows:
\begin{equation}\label{eq:predictive_mapping}
    \mathbf{v}_{i+1}^s = f(\mathbf{v}_i^s, \boldsymbol{\Phi}_i),
\end{equation}
where $\boldsymbol{\Phi}_i$ is the embedding of the aggregated information from $\mathcal{V}_{1:i}$ and $\mathcal{D}$ (detailed in Section~\ref{sec:catalyst_representation}), which plays a role of the catalyst during the generation. 

\smallskip
\noindent\underline{\textbf{Backward Denoising Diffusion Process.}}
The backward denoising diffusion process in existing diffusion models aims to reverse the above forward process and sample from $q(\mathbf{v}_i^{s-1}|\mathbf{v}_i^s)$ to recreate the true sample $\mathbf{v}_i^0$. Different from these approaches, our work aims to generate the next visit's representation, i.e., $\mathbf{v}_{i+1}^0$, using $\mathbf{v}_i^0$ based on their relationship modeled in Eq.~\eqref{eq:predictive_mapping}. 
Mathematically, the reverse process of $\mathbf{v}_{i+1}^0$ can be formulated as follows:
\begin{equation}
\begin{gathered}
    q(\mathbf{v}_{i+1}^{s-1}|\mathbf{v}_{i+1}^s,\mathbf{v}_{i+1}^0) = q(\mathbf{v}_{i+1}^s|\mathbf{v}_{i+1}^{s-1},\mathbf{v}_{i+1}^0)\frac{q(\mathbf{v}_{i+1}^{s-1}|\mathbf{v}_{i+1}^0)}{q(\mathbf{v}_{i+1}^s|\mathbf{v}_{i+1}^0)},\\
    q(\mathbf{v}_{i+1}^{s-1}|\mathbf{v}_{i+1}^s,\mathbf{v}_{i+1}^0) = \mathcal{N}(\mathbf{v}_{i+1}^{s-1};\hat{\boldsymbol{\mu}}_s(\mathbf{v}_{i+1}^s,\mathbf{v}_{i+1}^0),\hat{\beta}_s\mathbf{I}), 
\end{gathered}
\label{eq:backward}
\end{equation}

By simplifying Eq.~\eqref{eq:backward} according to the Gaussian distribution's density function, we can obtain the variance $\hat{\beta}_s$ and mean $\hat{\boldsymbol{\mu}}_s(\mathbf{v}_{i+1}^s,\mathbf{v}_{i+1}^0)$ of $q(\mathbf{v}_{i+1}^{s-1}|\mathbf{v}_{i+1}^s,\mathbf{v}_{i+1}^0)$ as follows:
\begin{equation}
    \begin{gathered}
        \hat{\beta}_s 
        =\frac{1-\bar{\alpha}_{s-1}}{1-\bar{\alpha}_s}\beta_s,\\
        \hat{\boldsymbol{\mu}}_s(\mathbf{v}_{i+1}^s,\mathbf{v}_{i+1}^0) 
        =\frac{\sqrt{\alpha}_s(1-\bar{\alpha}_{s-1})}{1-\bar{\mathbf{\alpha}_s}}  \mathbf{v}_{i+1}^s+ \frac{\sqrt{\bar{\alpha}_{s-1}}\beta_s}{1-\bar{\alpha}_s}\mathbf{v}_{i+1}^0.
    \end{gathered}
\label{eq:meanAndVar}
\end{equation}

Recall in the forward process, we have obtained $\mathbf{v}_{i+1}^s = \sqrt{\bar{\alpha}_s} \mathbf{v}_{i+1}^0 + \sqrt{1-\bar{\alpha}_s}\epsilon$ in Eq.~\eqref{eq:forwardRepa}. Thus, the mean value of the closed-form solution to the backward diffusion process can be obtained by substituting $\mathbf{v}_{i+1}^0$ in Eq.~\eqref{eq:meanAndVar} as follows:
\begin{equation}   \hat{\boldsymbol{\mu}}_s(\mathbf{v}_{i+1}^s,\mathbf{v}_{i+1}^0) = \frac{1}{\sqrt{\alpha}_s}(\mathbf{v}_{i+1}^s - \frac{1-\alpha_s}{\sqrt{1-\bar{\alpha}_s}}\epsilon_s).
\label{eq:meanAndVarofModel1}
\end{equation}

By substituting $\mathbf{v}_{i+1}^s$ in Eq.~\eqref{eq:meanAndVarofModel1} with the predictive mapping process in Eq.~\eqref{eq:predictive_mapping}, we finally have the closed-form solution as follows:
\begin{equation}   \hat{\boldsymbol{\mu}}_s(\mathbf{v}_{i+1}^s,\mathbf{v}_{i+1}^0) = \frac{1}{\sqrt{\alpha}_s}(f(\mathbf{v}_i^s, \boldsymbol{\Phi}_i) - \frac{1-\alpha_s}{\sqrt{1-\bar{\alpha}_s}}\epsilon_s).
\label{eq:mapped-meanAndVarofModel1}
\end{equation}

The details of the derivation of the backward reverse process can be found in Appendix Section~\ref{apd:backward}.

\noindent\underline{\textbf{\dmodel Learning.}}
We typically use a U-Net with parameters $\theta$ to train the proposed \dmodel by approximating Eq.~\eqref{eq:mapped-meanAndVarofModel1}, i.e.,
\begin{equation}
    \boldsymbol{\mu}_\theta^s(\mathbf{v}_{i+1}^s,s) = \frac{1}{\sqrt{\alpha}_t}(f(\mathbf{v}_i^s, \boldsymbol{\Phi}_i) - \frac{1-\alpha_t}{\sqrt{1-\bar{\alpha}_t}}\epsilon_\theta(f(\mathbf{v}_i^s, \boldsymbol{\Phi}_i),s)).
\label{eq:mapped-meanAndVarofModel2}
\end{equation}
Different from conventional U-Net architecture which only takes the noised embedding as input, we design a new predictive U-Net (\unet) equipped with the capability to condition on $\boldsymbol{\Phi}_i$ during the generation process. \unet also contains two paths of learning -- the downsampling path and the upsampling path. 

\unet takes $\mathbf{v}_{i}^s$ as the input of the first layer. Then, at each layer $l$, the downsampling operations include a ResNet block with a 1-D convolution operation to generate the input of layer $l+1$ as follows:
\begin{equation}
    \mathbf{v}_{i,l+1}^s = \text{Conv}(\text{ResBlock}(\mathbf{v}_{i,l}^s)).
\end{equation}

The upsampling path at the $l$-th layer consists of an information aggregator, a ResNet block, and a deconvolutional (DeConv) operation to reconstruct the input. The information aggregator is a self-attention block (SelfAtt) to fuse the embeddings of $\mathbf{v}_{i,l}^s$ and the transformed catalyst embedding $\boldsymbol{\Phi}_{i,l}$ by a linear function on $\boldsymbol{\Phi}_{i}$. The upsampling operation can be formulated as follows:
\begin{equation}
    \hat{\mathbf{v}}_{i+1, l}^s = \text{DeConv}(\text{ResBlock}(\hat{\mathbf{v}}_{i+1,l+1}^s, \text{SelfAtt}(\mathbf{v}_{i,l}^s, \boldsymbol{\Phi}_{i,l}))).
\label{selfatt}
\end{equation}

Figure~\ref{fig:unet} shows the designed \unet; the detailed derivation can be found in Appendix Section~\ref{apd:unet}. 

\begin{figure}[t]
    \centering
\includegraphics[width=0.85\columnwidth]{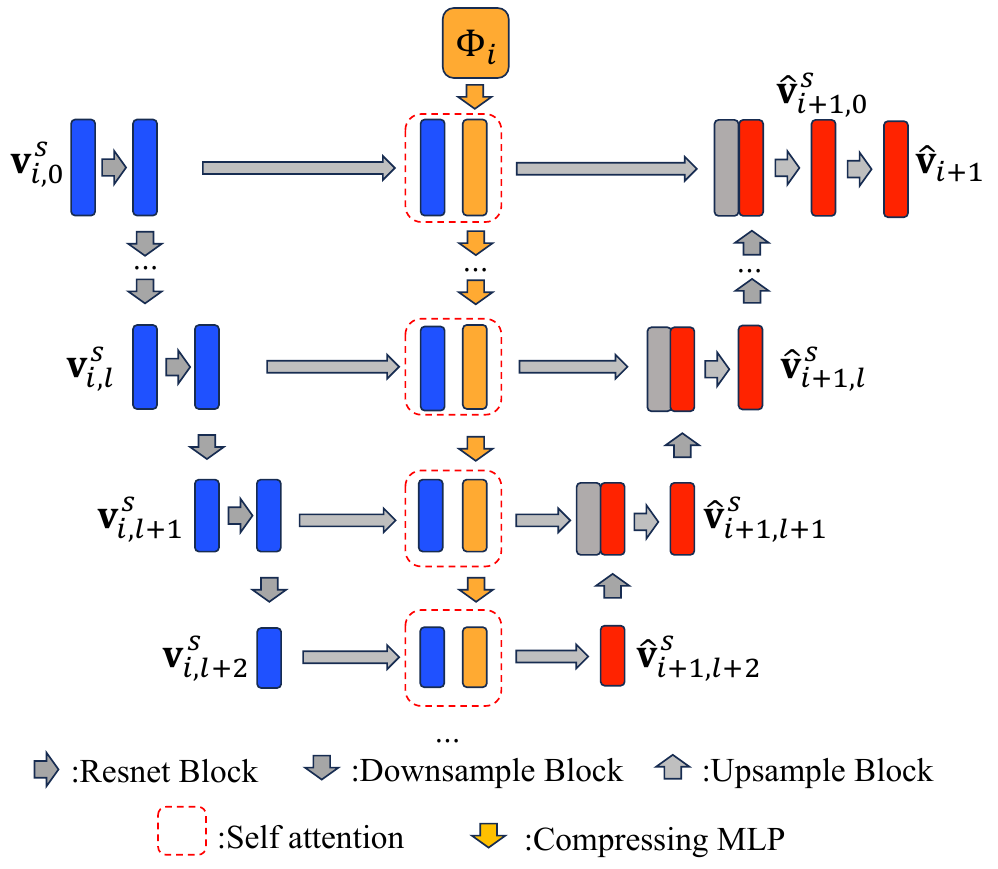}
    \vspace{-2mm}
    \caption{Illustration of \unet.}
    \label{fig:unet}
\vspace{-0.2in}
\end{figure}

\noindent\underline{\textbf{\dmodel Reconstruction Loss.}}
Let $\hat{\mathbf{v}}_{i+1} = \hat{\mathbf{v}}_{i+1, 0}^s$ denote the output of \unet that is trained on a randomly selected diffusion step $s \in [1, \cdots, S]$. The objective function of \unet is the mean squared errors between the generated $\hat{\mathbf{v}}_{i+1}$ and the learned embedding $\mathbf{v}_{i+1}$ in Section~\ref{sec:code-visit-embedding} at each training epoch as follows:
\begin{equation}\label{eq:diffusion_loss}
    \mathcal{L}_{d} (V_i) = \frac{1}{d_v} ||\hat{\mathbf{v}}_{i+1} - \mathbf{v}_{i+1}||^2,
\end{equation}
where $d_v$ is dimension size of $\hat{\mathbf{v}}_{i+1}$, and the learned visit embedding $\mathbf{v}_{i+1}$ can be treated as ground truths.

\noindent\underline{\textbf{Multimodal EHR Prediction.}}
The predicted embedding $\hat{\mathbf{v}}_{i+1}$ can also be used to predict medical codes in each modality. Specifically, for each modality $M^n_i$, we use a linear layer to map $\hat{\mathbf{v}}_{i+1}$ to a modality-level representation, and then a Sigmoid function is used to predict the probability of a medical code on top of a multilayer perceptron (MLP) as follows:
\begin{equation}
    \hat{\mathbf{y}}_{i+1}^{n} = \text{Sigmoid}(\text{MLP}_y( \hat{\mathbf{v}}_{i+1} )),
\end{equation}
Finally, we can use a Focal loss to train the multimodal EHR predictor as follows:
\begin{equation}\label{eq:ehr_prediction_loss}
\begin{split}
    \mathcal{L}_{e}(V_i) = &-\frac{1}{N}\sum_{n=1}^N \frac{1}{|\mathcal{C}_n|}\sum_{j=1}^{|\mathcal{C}_n|}[y_{i+1}^{n,j}\kappa(1-\hat{y}_{i+1}^{n,j})^\gamma\log(\hat{y}_{i+1}^{n,j})\\
    &+ (1-y_{i+1}^{n,j})(1-\kappa)(\hat{y}_{i+1}^{n,j})^\gamma\log(1-\hat{y}_{i+1}^{n,j})],
\end{split}
\end{equation}
where $|\mathcal{C}_n|$ denotes the number of identical codes in each modality $M_i^n$, $y_{i+1}^{n,j}$ is a binary ground truth to indicate whether the $j$-th code $c_{i+1}^{n,j}$ of the $n$-th modality presents in visit $V_{i+1}$, and $\kappa$ and $\gamma$ are hyperparameters.






To optimize Eqs.~\eqref{eq:diffusion_loss} and~\eqref{eq:ehr_prediction_loss}, we need to obtain the catalyst representation $\boldsymbol{\Phi}_i$. Next, we introduce the details of catalyst representation learning in Section~\ref{sec:catalyst_representation}.

\subsection{Catalyst Representation $\boldsymbol{\Phi}_i$ Learning }~\label{sec:catalyst_representation}
As discussed in Section~\ref{sec:pddpm}, the catalyst information is significantly important in the proposed \ours during the generation with \dmodel, which ``translates'' the information from $\mathbf{v}_i^s$ to $\mathbf{v}_{i+1}^s$ at each diffusion step $s$ via Eq.~\eqref{eq:predictive_mapping}. Next, we explain how we construct the catalyst representation $\boldsymbol{\Phi}_i$.

\noindent\underline{\textbf{EHR Historical Information Representation.}}
In clinical practice, professionals often rely on a patient's historical medical records for a comprehensive view of their past health issues and as a crucial tool for informed decision-making. These records offer a timeline of medical events, treatments, and diagnoses that provide insights into the patient's health trajectory and are useful for predicting future health scenarios. Thus, we incorporate historical medical information as one of the conditioning factors to aid our generation process. Based on the learned visit embedding using Eq.~\eqref{eq:visit_embedding}, we utilize an LSTM network to accumulate a hidden state $\mathbf{h}_i$ for each visit as follows:
\begin{equation}\label{eq:lstm}
    \mathbf{h}_i = \text{LSTM}(\mathbf{h}_{i-1},\mathbf{v}_i).
\end{equation}

\noindent\underline{\textbf{Time Interval Estimation.}}
Not only do clinical professionals diagnose a patient's health condition, but they also make a crucial decision in determining the optimal timing for the follow-up visit that best suits the current health condition of the patient. This decision is often to cope with the urgency and nature of the patient's condition: patients suffering from acute illnesses may need to return within a matter of days, while those with chronic diseases revisit with a more prolonged and periodic pattern. In our approach, we use the current health condition $\mathbf{h}_i$ to make predictions on the time interval till the next follow-up visit with a continuous time LSTM~\cite{hawkeslstm}, as shown in Eq.~\eqref{eq:timepred}. We utilize a linear layer on hidden state $\mathbf{h}_i$ to learn an intensity measure $\boldsymbol{\lambda}_i$ of the current visit, which represents a patient's medical urgency. This intensity is subtracted from $1$, giving a close to $0$ output if the patient's condition is urgent. Then, the second equation predicts the time gap and ensures it is strictly above $0$ with the Softplus activation function:
\begin{equation}
\begin{aligned}
    \boldsymbol{\lambda}_i &= 1 - \text{tanh}(\text{MLP}_{\lambda}(\mathbf{h}_i)),\\
    \hat{\Delta}_i &= \text{Softplus}(\text{MLP}_{\Delta}(\boldsymbol{\lambda}_i)).
\end{aligned}
\label{eq:timepred}
\end{equation}

\noindent\underline{\textbf{Demographic Information Embedding.}}
Demographic information is also treated as a key factor in decision-making. Thus, we encode the demographic information $\mathcal{D}$ into a dense representation $\mathbf{d}$ using an MLP layer, i.e., $\mathbf{d} = \text{MLP}_d (\mathcal{D})$.

\noindent\underline{\textbf{Catalyst Representation Learning.}}
Finally, the catalyst representation $\boldsymbol{\Phi}_i = [\mathbf{h}_i; \hat{\boldsymbol{\Delta}}_i; \mathbf{d}]$ is obtained by concatenating the historical representation $\mathbf{h}_i$, the embedding of time interval  through the positional embedding on $\hat{{\Delta}}_i$ (i.e., $\hat{\boldsymbol{\Delta}}_i$), and the demographic embedding $\mathbf{d}$.

\subsection{Time Interval Estimation Loss}
The proposed \ours generates not only the next visit information but also the time interval between visits $V_i$ and $V_{i+1}$.
We take another MSE loss $\mathcal{L}_\text{time}$ between the real-time gap ${\Delta}_i$ and the predicted time gap $\hat{\Delta}_i$ using Eq.~\eqref{eq:timepred} as follows:
\begin{equation}
\begin{gathered}
\mathcal{L}_t(V_i) = ({\Delta}_i - \hat{\Delta}_i)^2,
\end{gathered}
\label{timeLoss}
\end{equation}

\subsection{\ours Loss}
Finally, we define the total loss $\mathcal{L}$ of a patient with $|\mathcal{V}|$ visits as the weighted sum of all three loss components by $\omega_{d}$, $\omega_e$, and $\omega_t$, as follows:
\begin{equation}\label{eq:mainloss}
    \mathcal{L} = \frac{1}{|\mathcal{V}|-1}\sum_{i=1}^{|\mathcal{V}|-1}\left(\omega_d \mathcal{L}_{d}(V_i) + \omega_e \mathcal{L}_{e}(V_i) + \omega_t \mathcal{L}_{t}(V_i)\right).
\end{equation}

\section{Experiment}


Due to the limited space, we put more results in the Appendix.

\subsection{Datasets}
We use two publicly available datasets to validate the performance of the proposed \ours, including MIMIC-III~\cite{mimiciii} and Breast Cancer Trial\footnote{Clinical Trial ID: NCT00174655, \url{https://www.projectdatasphere.org/}} from Project Data Sphere. The statistics of these two datasets are listed in Table~\ref{table:dataset}.
For the \textbf{MIMIC-III} dataset, we extract all $46,520$ patients' diagnoses, prescriptions, lab items, and procedure codes as four modalities of interest. 
For the \textbf{Breast Cancer Trial} dataset, we mainly follow the data preprocessing procedure described in TWIN~\cite{twin}, extracting adverse events, medications, lab categories, and treatment codes. For both datasets, each patient's EHR data is represented by a sequence of admissions ordered by admission time, where each admission consists of four lists of codes from each modality accordingly. Lastly, we add demographic information, such as sex, age, race, etc., as a static feature vector. We randomly split each dataset into train, validation, and test sets, with a ratio of $75:10:15$.

\subsection{Implementation Details}

Our model is implemented in PyTorch and trained on an NVIDIA RTX A6000 GPU. We use the Adam optimizer with learning rate and weight decay both set to \(10^{-3}\). We set the Focal Loss parameter in Eq.~\eqref{eq:ehr_prediction_loss} to $\kappa = 0.75$ and $\gamma = 5$. For the total \ours loss in Eq.~\eqref{eq:mainloss}, we set $\omega_d = 0.5$, $\omega_e = 1000$, and $\omega_t = 0.01$. The dimension of $\mathbf{h}$, $\mathbf{v}$, $\mathbf{\Delta}$, and $\mathbf{d}$ are set to $256$, and the \unet dimension list is $[1024,512,256]$.


\begin{table}[t]
{
\begin{tabular}{c|c||c|c}
\toprule
\multicolumn{2}{c||}{MIMIC-III}  & \multicolumn{2}{c}{Breast Cancer Trial} \\ \hline
Total Patients          & 46,520      & Total Patients &   970              \\
Diagnosis  &    1,071     & Adverse Events &     50        \\
Drug Codes   &    1,439    & Medications &  100  \\
Lab Items      &    710      & Lab Categories & 9             \\
Procedure Codes      &    711   & Treatments &  4 \\
\bottomrule
\end{tabular}
}
\caption{Statistics of two main datasets.}
\label{table:dataset}
\vspace{-0.3in}
\end{table}

\begin{table*}[]
\centering
\resizebox{1\textwidth}{!}{
\begin{tabular}{c|c|c|c|c|c|c|c|c|c|c|c|c|c}
\toprule

\textbf{Dataset} & \textbf{Modality} & \textbf{Metric} & MLP    & medGAN & synTEG & EVA   & TWIN  & TabDDPM & Meddiff & ScoEHR & PromptEHR     & HALO   & \ours         \\ \hline
\multirow{8}{*}{\rotatebox{90}{MIMIC-III}}  & \multirow{2}{*}{Diagnosis}    & LPL             & 325.55 & 242.30 & 36.87  & 29.62 & 26.28 & 108.79  & 664.54  & 685.17 & 126.23        & 149.66 & \textbf{15.97} \\
                 &                   & MPL             & 352.54 & 257.48 & 45.61  & 31.63 & 27.68 & 114.18  & 670.91  & 691.55 & 128.05        & 192.13 & \textbf{17.95} \\ \cline{2-14} 
                 & \multirow{2}{*}{Drug}              & LPL             & 553.63 & 403.02 & 83.38  & 43.79 & 40.94 & 179.22  & 936.28  & 934.87 & 167.48        & 166.11 & \textbf{20.53} \\
                 &                   & MPL             & 551.67 & 405.75 & 82.66  & 44.02 & 40.86 & 178.70  & 936.14  & 950.27 & 136.04        & 202.01 & \textbf{19.15} \\ \cline{2-14} 
                 & \multirow{2}{*}{Lab Item}          & LPL             & 168.25 & 77.10  & 26.80  & 20.05 & 17.47 & 54.69   & 413.41  & 432.11 & 107.22        & 322.51 & \textbf{15.11} \\
                 &                   & MPL             & 166.61 & 87.12  & 30.34  & 19.97 & 17.41 & 54.44   & 412.33  & 431.09 & 98.52         & 303.09 & \textbf{13.99} \\ \cline{2-14} 
                 & \multirow{2}{*}{Procedure}         & LPL             & 290.38 & 234.81 & 49.33  & 27.39 & 21.26 & 98.03   & 471.81  & 486.81 & 51.18         & 22.68  & \textbf{14.53} \\
                 &                   & MPL             & 286.53 & 245.28 & 44.00  & 30.49 & 24.26 & 102.72  & 479.96  & 499.89 & 31.13         & 39.04  & \textbf{18.89} \\ \hline
\multirow{8}{*}{\rotatebox{90}{Breast Cancer Trial}} & \multirow{2}{*}{Adverse Event} & LPL             & 8.42   & 8.00   & 8.21   & 6.08  & 6.08  & 9.31    & 49.04   & 51.82  & 12.37         & 34.83  & \textbf{5.96}  \\
                 &                   & MPL             & 9.37   & 9.42   & 9.70   & 8.30  & 8.52  & 10.86   & 50.13   & 51.57  & 12.14         & 31.51  & \textbf{8.02}  \\ \cline{2-14} 
                 &\multirow{2}{*}{Medication}        & LPL             & 8.82   & 9.53   & 8.21   & 5.39  & 5.56  & 11.13   & 99.35   & 99.31  & 19.34         & 31.22  & \textbf{4.96}  \\
                 &                   & MPL             & 8.73   & 11.67  & 10.08  & 6.95  & 7.10  & 12.95   & 98.83   & 99.10  & 19.80         & 33.61  & \textbf{5.87}  \\ \cline{2-14} 
                 & \multirow{2}{*}{Lab Category}      & LPL             & 9.33   & 10.41  & 9.63   & 9.07  & 9.09  & 9.06    & 10.95   & 10.93  & \textbf{8.55} & 9.14   & 9.01           \\
                 &                   & MPL             & 9.22   & 10.08  & 10.03  & 9.09  & 9.11  & 9.03    & 10.96   & 10.97  & \textbf{8.66} & 9.28   & 9.09           \\ \cline{2-14} 
                 & \multirow{2}{*}{Treatment}         & LPL             & 7.29   & 9.43   & 9.09   & 3.09  & 3.12  & 3.67    & 4.77    & 5.01   & 5.10          & 3.44   & \textbf{2.63}  \\
                 &                   & MPL             & 4.47   & 4.83   & 4.43   & 2.89  & 2.92  & 3.22    & 4.84    & 5.00   & 5.63          & 3.05   & \textbf{2.41}  \\
\bottomrule
\end{tabular}
}
\caption{EHR data generation evaluation of different approaches on two datasets with two metrics.}
\label{table:lplmpl}
\vspace{-0.2in}
\end{table*}

\subsection{Fidality Assessment}

We perform experiments to evaluate the generated data quality with two evaluation metrics and various baseline models, emphasizing the temporal coherence and cross-modality consistency of the generated data.

\subsubsection{Baselines}

Our selected baseline models include MLP, GAN-based models (medGAN~\cite{medGAN} and synTEG~\cite{synteg}), VAE-based models (EVA~\cite{eva} and TWIN~\cite{twin}), diffusion-based approaches (TabDDPM~\cite{tabddpm}, Meddiff~\cite{he2023meddiff}, and ScoEHR~\cite{naseer2023scoehr}), and language model-based approaches (PromptEHR~\cite{promptehr} and HALO~\cite{HALO}). Appendix Section~\ref{generationBaseline} describes each model's detailed explanation.

\subsubsection{Experiment Design and Evaluation Metrics}

In this experiment, we use MIMIC-III and the Breast Cancer Trial as input databases separately. Each model produces a synthetic dataset corresponding to the original one, maintaining a 1:1 ratio. To assess the effectiveness of these EHR generation models, we focus on the following evaluation metrics. \textbf{Longitudinal Imputation Perplexity (LPL)} is a specialized metric used to evaluate EHR generation models. This metric adapts the traditional concept of perplexity from natural language processing to suit the unique temporal structure of EHR data. The LPL metric effectively captures the model's ability to predict the sequence of medical events over time, considering the chronological progression of a patient's health condition. In contrast to the LPL, which concentrates on the temporal coherence within a single modality, \textbf{Cross-modality Imputation Perplexity (MPL)} extends this concept to encompass the interrelations among different modalities, by assessing the model's proficiency in integrating and predicting across various types of data modalities, making it a more comprehensive measure of a model's ability to handle the multifaceted nature of EHR data.

\subsubsection{Experimental Results}
Table~\ref{table:lplmpl} shows the experimental results on the LPL and MPL metrics of all models tested on each of the data sources. \textit{Lower score, indicates better model performance}. On the MIMIC-III dataset, our proposed model consistently outperforms other models across all four modalities in both LPL and MPL metrics. For instance, in the Diagnosis modality, our model achieves the best LPL score of 15.97 and MPL score of 17.95, significantly better than the next best baseline model, TWIN, which scored 26.28 and 27.68 in LPL and MPL, respectively. Though PromptEHR slightly outperforms our model in the Lab Category modality within the Breast Cancer Trial dataset, its performance across other modalities is less consistent. This variation indicates that while PromptEHR can be effective in certain scenarios, its output generally exhibits greater variability and less reliability compared to our model. Such inconsistency can lead to diminished effectiveness in diverse medical data scenarios, underlining our model's superior adaptability and robustness. Overall, our model's consistent performance across various metrics and modalities reinforces its effectiveness and broad applicability in medical data generation.

\begin{table}[t]
{
\begin{tabular}{c|cccc}
\toprule
Approach  & 10\% & 20\% & 35\% & 50\%  \\ \hline
MLP          & 13.53 & 13.38 & 13.03 & 13.07 \\
medGAN       & 17.06 & 17.19 & 17.56 & 17.79 \\
synTEG       & 13.21 & 13.02 & 12.71 & 12.76 \\
EVA          & 13.36 & 13.17 & 12.84 & 13.36 \\
TWIN         & 13.36 & 13.16 & 12.84 & 12.89 \\
TabDDPM      & 13.45 & 13.66 & 13.50 & 13.68 \\
Meddiff      & 14.94 & 17.00 & 18.35 & 19.18 \\
ScoEHR       & 14.12 & 15.56 & 16.44 & 16.91 \\
PromptEHR    & 14.44 & 12.86 & 12.90 & 13.31 \\
HALO         & 13.52 & 13.79 & 13.88 & 13.79 \\ \hline
\ours         & \textbf{12.60} & \textbf{12.77} & \textbf{12.53} & \textbf{12.25} \\
\bottomrule
\end{tabular}
}
\caption{Privacy assessment on MIMIC-III with different percentages of known patients under the metric PD.}
\label{table:pd}
\vspace{-0.3in}
\end{table}

\subsection{Privacy Assessment}

We also evaluate the privacy-preserving capability of our model against other generation baseline models. The privacy-preserving capability is how likely the generated data can be traced back to the original data. We conduct our experiments with the Presence Disclosure Sensitivity metric.

\subsubsection{Experiment Design and Evaluation Metric}
We start with a predefined percentage of patient records from the training set, labeling them as ``\textit{known}'' or ``\textit{compromised}''. The aim is to identify these known records within the generated dataset. If the $i$-th visit of a patient is matched back to one of the synthetic visits generated by this patient by similarity score, we count it as a successful attack. We use the metric \textbf{Presence Disclosure Sensitivity (PD)}~\cite{twin} to evaluate the security of our datasets. PD is the proportion of known patient records correctly matched in the generated dataset against the total number of compromised records. \textit{The lower the PD value, the better the security performance}. This metric effectively gauges the risk of individual patient identification in the generated dataset, serving as an indicator of the dataset's privacy and data protection capabilities.


\subsubsection{Experimental Results}
Table~\ref{table:pd} shows the experiment results on MIMIC-III in terms of PD with varying percentages of known patient records, ranging from 10\% to 50\%. Our analysis reveals that our model consistently outperforms the baseline models across all tested scenarios. Notably, as the percentage of known patients increases, our model maintains its effectiveness in protecting patient privacy. For instance, at 10\% known patients, our model achieves a Presence Disclosure Sensitivity of 12.60, and even with 50\% known patients, it has the lowest metric of 12.25. This demonstrates a robust defense against privacy breaches, even as the challenge escalates with more known patient records. In comparison, other models like medGAN, TabDDPM, and PromptEHR show higher sensitivity, indicating a greater risk of patient identification in their generated datasets. For example, medGAN's sensitivity ranges from 17.06 to 17.79, which is significantly worse than ours. These results underscore the effectiveness of our model in ensuring the privacy and protection of patient data.

\begin{table*}[t]
{
\begin{tabular}{l|cccccccccccc}
\toprule

\textbf{Model }         & \multicolumn{3}{c|}{F-LSTM}         & \multicolumn{3}{c|}{F-CNN}              & \multicolumn{3}{c|}{RAIM}               & \multicolumn{3}{c}{DCMN} \\ \hline
\textbf{ Metric}  & AUPR & F1 & \multicolumn{1}{c|}{Kappa} & AUPR & F1 & \multicolumn{1}{c|}{Kappa} & AUPR & F1 & \multicolumn{1}{c|}{Kappa} & AUPR     & F1     & Kappa     \\ \hline

Orginal       &0.5710&0.4705& \multicolumn{1}{l|}{0.4221}      &0.5810&0.5132& \multicolumn{1}{l|}{0.4554}      &0.5849&0.5000& \multicolumn{1}{l|}{0.4280}      &0.5438&0.4742&0.4298  \\
MLP      &0.6344&0.5408& \multicolumn{1}{l|}{0.4747}      &0.6614&0.5882& \multicolumn{1}{l|}{0.4950}      &0.6226&0.5571& \multicolumn{1}{l|}{0.4819}      &0.5733&0.4975&0.4245  \\
medGAN   &0.6210&0.5685& \multicolumn{1}{l|}{0.4946}      &0.6563&0.6098& \multicolumn{1}{l|}{0.5337}      &0.6159&0.5455& \multicolumn{1}{l|}{0.4789}      &0.5668&0.5473&0.4793  \\
synTEG   &0.6309&0.5556& \multicolumn{1}{l|}{0.4815}      &0.6597&0.6026& \multicolumn{1}{l|}{0.5220}      &0.6490&0.5891& \multicolumn{1}{l|}{0.5185}      &0.5804&0.5674&0.4958  \\
EVA      &0.6313&0.5572& \multicolumn{1}{l|}{0.4907}      &0.6487&0.5703& \multicolumn{1}{l|}{0.4897}      &0.6366&0.5438& \multicolumn{1}{l|}{0.4890}      &0.5585&0.5076&0.4326  \\
TWIN     &0.6410&0.5503& \multicolumn{1}{l|}{0.4846}      &0.6642&0.5929& \multicolumn{1}{l|}{0.5412}      &0.6469&0.5876& \multicolumn{1}{l|}{0.5292}      &0.6283&0.5687&0.4992  \\ 
TabDDPM  &0.6489&0.5586& \multicolumn{1}{l|}{0.4939}      &0.6534&0.5672& \multicolumn{1}{l|}{0.5022}      &0.6228&0.5572& \multicolumn{1}{l|}{0.4858}      &0.5428&0.5112&0.4354  \\
Meddiff  &0.6337 & 0.5502 & \multicolumn{1}{l|}{0.4823}      & 0.6163 & 0.5504 & \multicolumn{1}{l|}{0.4636}      & 0.6161 & 0.5289 & \multicolumn{1}{l|}{0.4597}      & 0.5594 & 0.4969 & 0.4186 \\
ScoEHR   &0.6408 & 0.5438 & \multicolumn{1}{l|}{0.4812}      & 0.6392 & 0.6033 & \multicolumn{1}{l|}{0.5272}      & 0.5949 & 0.4964 & \multicolumn{1}{l|}{0.4191}      & 0.6089 & 0.5333 & 0.4594 \\
PromptEHR&0.6580&0.5677& \multicolumn{1}{l|}{0.5041}      &0.6682&0.6079& \multicolumn{1}{l|}{0.5383}      &0.6419&0.5648& \multicolumn{1}{l|}{0.4923}      &0.6279&0.6036&0.5351  \\
HALO     &\textbf{0.6673}&0.5547& \multicolumn{1}{l|}{0.4885}      &0.6139&0.5234& \multicolumn{1}{l|}{0.4562}      &0.5812&0.4779& \multicolumn{1}{l|}{0.4119}      &0.6124&0.5746&0.4957  \\\hline
\ours   &0.6658&\textbf{0.5870}& \multicolumn{1}{l|}{\textbf{0.5251}}      &\textbf{0.6835}&\textbf{0.6159}& \multicolumn{1}{l|}{\textbf{0.5425}}      &\textbf{0.6548}&\textbf{0.5936}& \multicolumn{1}{l|}{\textbf{0.5327}}      &\textbf{0.6385}&\textbf{0.6147}&\textbf{0.5448}  \\ 
\bottomrule
\end{tabular}
}
\caption{Result evaluation of the Mortality task on multimodal EHR data.}
\label{table:multimodal}
\vspace{-0.15in}
\end{table*}

\begin{table*}[t]
\resizebox{1\textwidth}{!}
{
\begin{tabular}{l|ccccccccccccccc}
\toprule
\textbf{Backbone  }        & \multicolumn{3}{c|}{Adacare}         & \multicolumn{3}{c|}{Dipole}     & \multicolumn{3}{c|}{HiTANet}      & \multicolumn{3}{c|}{LSTM}  & \multicolumn{3}{c}{Retain}\\ \hline
\textbf{Metric}  & AUPR & F1 & \multicolumn{1}{l|}{Kappa} & AUPR & F1 & \multicolumn{1}{l|}{Kappa} & AUPR & F1 & \multicolumn{1}{l|}{Kappa} & AUPR & F1 & \multicolumn{1}{l|}{Kappa} & AUPR     & F1     & Kappa     \\ \hline
Orginal           &0.6242&0.6136& \multicolumn{1}{l|}{0.3627} &0.5856&0.5740& \multicolumn{1}{l|}{0.3349} &0.6203&0.5978& \multicolumn{1}{l|}{0.3740} &0.5943&0.5758& \multicolumn{1}{l|}{0.3461} &0.5989&0.5913& \multicolumn{1}{l}{0.3720}\\
MLP               &0.6640&0.6376& \multicolumn{1}{l|}{0.4185} &0.6872&0.6470& \multicolumn{1}{l|}{0.4511} &0.6692&0.6562& \multicolumn{1}{l|}{0.4488} &0.6706&0.6374& \multicolumn{1}{l|}{0.4460} &0.6476&0.6279& \multicolumn{1}{l}{0.4158}\\
medGAN            &0.6669&0.6400& \multicolumn{1}{l|}{0.4089} &0.6915&0.6371& \multicolumn{1}{l|}{0.4490} &0.6781&0.6492& \multicolumn{1}{l|}{0.4376} &0.6667&0.6332& \multicolumn{1}{l|}{0.4431} &0.6424&0.6285& \multicolumn{1}{l}{0.4028}\\
synTEG            &0.6711&0.6121& \multicolumn{1}{l|}{0.4142} &0.6820&0.6215& \multicolumn{1}{l|}{0.4174} &0.6851&0.6607& \multicolumn{1}{l|}{0.4641} &0.6676&0.6312& \multicolumn{1}{l|}{0.4353} &0.6319&0.6285& \multicolumn{1}{l}{0.4224}\\
EVA               &0.6590&0.6424& \multicolumn{1}{l|}{0.4189} &0.6795&0.6527& \multicolumn{1}{l|}{0.4513} &0.6813&0.6511& \multicolumn{1}{l|}{0.4372} &0.6629&0.6307& \multicolumn{1}{l|}{0.4267} &0.6527&0.6240& \multicolumn{1}{l}{0.4208}\\
TWIN              &0.6739&0.6252& \multicolumn{1}{l|}{0.4328} &0.6603&0.6409& \multicolumn{1}{l|}{0.4406} &0.6789&0.6690& \multicolumn{1}{l|}{0.4274} &0.6546&0.6264& \multicolumn{1}{l|}{0.4162} &0.6540&0.6382& \multicolumn{1}{l}{0.4187}\\
TabDDPM           &0.6677&0.6243& \multicolumn{1}{l|}{0.3877} &0.6851&0.6342& \multicolumn{1}{l|}{0.4312} &0.6633&0.6581& \multicolumn{1}{l|}{0.4480} &0.6687&0.6317& \multicolumn{1}{l|}{0.4301} &0.6465&0.6152& \multicolumn{1}{l}{0.4067}\\
Meddiff           &0.6659&0.6323& \multicolumn{1}{l|}{0.4171} &0.6756&0.6249& \multicolumn{1}{l|}{0.4190} &0.6684&0.6301& \multicolumn{1}{l|}{0.4277} &0.6672&0.6188& \multicolumn{1}{l|}{0.4119} &0.6591&0.6204& \multicolumn{1}{l}{0.4161}\\
ScoEHR            &0.6701&0.6395& \multicolumn{1}{l|}{0.4284} &0.6719&0.6296& \multicolumn{1}{l|}{0.4117} &0.6774&0.6340& \multicolumn{1}{l|}{0.4238} &0.6624&0.5980& \multicolumn{1}{l|}{0.3966} &0.6469&0.6282& \multicolumn{1}{l}{0.4166}\\
PromptEHR         &0.6810&0.6462& \multicolumn{1}{l|}{0.4100} &0.6748&0.6359& \multicolumn{1}{l|}{0.4334} &0.6541&0.6182& \multicolumn{1}{l|}{0.4076} &0.6642&0.6222& \multicolumn{1}{l|}{0.4178} &0.6582&0.6251& \multicolumn{1}{l}{0.4211}\\
HALO              &0.6742&0.6312& \multicolumn{1}{l|}{0.4295} &0.6907&0.6562& \multicolumn{1}{l|}{0.4604} &0.6841&0.6578& \multicolumn{1}{l|}{0.4489} &0.6619&0.6301& \multicolumn{1}{l|}{0.4252} &0.6518&0.6266& \multicolumn{1}{l}{0.4196}\\\hline
\ours             &\textbf{0.6856}&\textbf{0.6523} & \multicolumn{1}{l|}{\textbf{0.4385}} &\textbf{0.7018}&\textbf{0.6630}& \multicolumn{1}{l|}{\textbf{0.4735}}  &\textbf{0.7017}&\textbf{0.6777}& \multicolumn{1}{l|}{\textbf{0.4699}} &\textbf{0.6824}&\textbf{0.6484}& \multicolumn{1}{l|}{\textbf{0.4506}} &\textbf{0.6603}&\textbf{0.6397}& \multicolumn{1}{l}{\textbf{0.4382}}\\ 
\bottomrule
\end{tabular}
}
\caption{Result evaluation on Heart Failure prediction task on unimodal EHR data.}
\label{table:unimodal}
\vspace{-0.2in}
\end{table*}

\subsection{Utility Assessment}

We experiment with the utility of the generated dataset from two databases on various downstream tasks under both multimodal and unimodal settings. We also conduct experiments to assess the effectiveness of the time gap prediction module of our model.

\subsubsection{Data Preprocessing}
We follow the FIDDLE~\cite{flstmandfcnn} guidelines for data preprocessing and adapt their label definitions to process the MIMIC-III database, focusing on three critical health outcomes of a multimodal setting: Acute Respiratory Failure (ARF), Shock, and Mortality. Additionally, we employ another data preprocessing method from Retain~\cite{retain} to obtain diagnosis codes for heart failure risk prediction, demonstrating our model's effectiveness in an unimodal context. Furthermore, following the work of TWIN~\cite{twin}, we select patients with severe outcomes and death as positive labels.

\subsubsection{Multimodal Risk Prediction Analysis}\label{multiRiskPred}

To evaluate the quality of synthetic data generated by our approach, we designed an experiment to determine whether integrating synthetic data into the training process enhances the performance of downstream task-oriented models. We take all four time-series modalities and stationary demographic information as input features to conduct multimodal risk prediction experiments on acute respiratory failure (ARF), Shock, and Mortality datasets. We choose the following models as baselines: F-LSTM~\cite{flstmandfcnn}, F-CNN~\cite{flstmandfcnn}, RAIM~\cite{raim}, and DCMN~\cite{dcmn}, and three evaluation metrics: AUPR (the area under the Precision-Recall curve), F1 and Kappa, following~\cite{medhmp}. \textit{Baseline models and the results on ARF and Shock are explained in Appendix Section~\ref{apd:multimodalPrediction} and Section~\ref{apd:multiRiskPred}, respectively.} 

For each dataset, models are trained using either only original data or a blend of synthetic and original data at a 1:1 ratio. The results, detailed in Table~\ref{table:multimodal}, reveal that our method, \ours, consistently outperforms baseline models under most conditions. Notably, under the Mortality task with the F-CNN architecture, \ours demonstrates a 2\% improvement in both AUPR and F1 metrics compared to baselines. In contrast, the HALO model shows superior performance in specific metrics when paired with the F-LSTM architecture, suggesting a particularly effective synergy between HALO's synthetic data and the F-LSTM model. Overall, these results imply our model's potential to provide reliable synthetic data to augment multimodal risk prediction models.

\subsubsection{Unimodal Risk Prediction Analysis}\label{uniRiskPred}

To simulate a scenario where multimodal data are unavailable, we conduct the following unimodal risk prediction task on Heart Failure disease with diagnosis code only. The backbone risk prediction models are LSTM~\cite{lstm}, Dipole~\cite{dipole}, Retain~\cite{retain}, AdaCare~\cite{adacare}, and HiTANet~\cite{hitanet}, and are explained in Appendix Section~\ref{apd:unimodalPrediction}. We utilize the same evaluation metric and synthetic-real data ratio as the multimodal experiment. 

The results of these experiments in Table~\ref{table:unimodal} show our model outperforms other generation models. Compared to the best-performing model PromptEHR with Adacare, our model achieves a 3\% higher Kappa. One notable point is that to make a fair comparison between baseline models, the real data does not contain the time interval between visits. Thus, backbone methods that rely on learning time information, such as HiTANet, do not perform optimally. However, we can see that when our model provides extra time information in the synthetic dataset, HiTANet's performance greatly increases and is better than others: the AUPR and F1 of HiTANet rise by 8\%. This experiment not only underscores our model's effectiveness in generating high-quality unimodal data but also demonstrates its unique capability to enrich synthetic datasets with critical temporal information, thereby offering comprehensive support for advanced predictive analytics.

\subsubsection{Time Interval Prediction}\label{timePred}

While the previous experiment implicitly confirmed our model's time interval prediction capability and effectiveness for downstream risk prediction, we directly compare ours to a range of established time-series forecasting baselines in this experiment. The task is defined to use historical time stamps till ${T}_i$ to predict $T_{i+1}$ and will include ${V_{i}}$ if the model structure allows. The selected baseline methods include the Autoregressive Integrated Moving Average (ARIMA)~\cite{arima}, Support Vector Regression (SVR)~\cite{svr}, Gradient Boosting Regression Trees (GBRT)~\cite{gbrt}, Kalman Filter (KF)~\cite{kalman}, and LSTM~\cite{lstm}, which are explained in Appendix Section~\ref{apd:timePrediction}. We use Root Mean Squared Error (RMSE) as the metric. The less the RMSE, the better the fit of the model. 

\begin{figure}
    \centering
    \includegraphics[width=0.9\columnwidth]{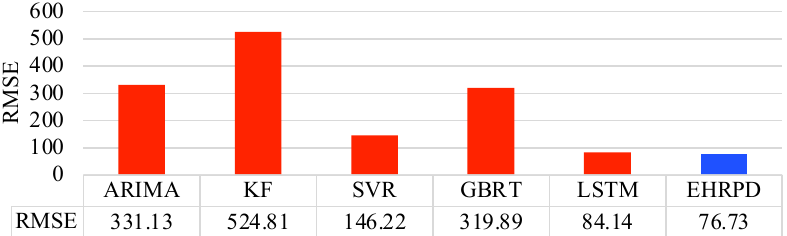}
    \vspace{-0.1in}
    \caption{Illustration of time interval prediction with RMSE.}
    \label{fig:timeplot}
    \vspace{-0.2in}
\end{figure}

We visualize the results in Figure~\ref{fig:timeplot}, with red-colored bars representing baseline models and blue representing ours, as well as RMSE values on the bottom. We can see that timestamp-only methods do not perform well. The conventional LSTM performs closely to \ours, while \ours shows the best performance with the lowest RMSE. This experiment demonstrates our model's ability to integrate various additional information and deliver more accurate predictions on the intervals leading up to the next patient visit.

\subsubsection{Severe Outcome Prediction}

In the healthcare domain, datasets often have a smaller scale compared to extensive public resources like MIMIC-III, highlighting the necessity for generation models to efficiently generate with limited data. With this concern, we assess our method's performance on the Breast Cancer Trial dataset, which presents a challenging environment with relatively few data entries. Following the settings in TWIN~\cite{twin}, our task is to predict the severe outcome and death defined in the data preprocessing section, and the selected metric is the Area Under the Receiver Operating Characteristic (AUROC) score. The size of the generated dataset is equal to the training dataset. 

\begin{figure}
    \centering
    \includegraphics[width=0.95\columnwidth]{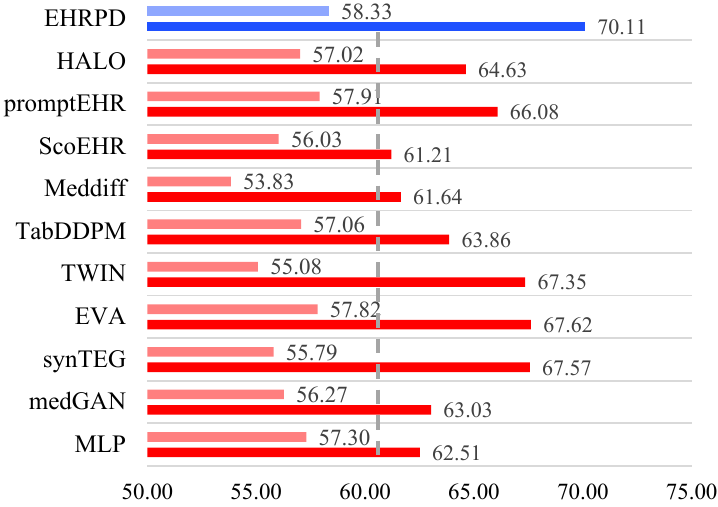}
    \vspace{-0.1in}
    \caption{Illustration of severe outcome prediction from the synthetic or synthetic-real hybrid datasets.}
    \label{fig:seplot}
    \vspace{-0.2in}
\end{figure}

An LSTM network is used to learn the sequential visit-level hidden states, which are then utilized by an MLP to make a binary prediction. The results are depicted in Figure~\ref{fig:seplot}. The dashed line represents the AUROC value achieved using the real dataset. Our model is colored in blue, while baselines are colored in red. For each pair of the histogram, the light-colored one is the performance from synthetic data only, while the darker one is from synthetic data plus real data. Our model achieves the best performance under both synthetic-only and hybrid settings. This comparative experiment provides a clear visualization of our model's capability to generate synthetic data that is both realistic and effective for advanced predictive tasks.

\subsubsection{Adverse Event Prediction}

While the previous experiments show our model's generation capability on the patient level, now we evaluate the fine-grained code-level generation capability and see whether the generated visit is coherent with its predecessor. Thus, in this section, our task is to predict the next visit's adverse events with the current visit's multimodal codes. The only training datasets available are the synthetic ones with AUROC as the metric. The size of the generated dataset is equal to the real training dataset. We utilize linear layers to embed medical codes and aggregate to visit level, and then an MLP predicts the next visit's adverse events. 

\begin{figure}
    \centering
    \includegraphics[width=0.9\columnwidth]{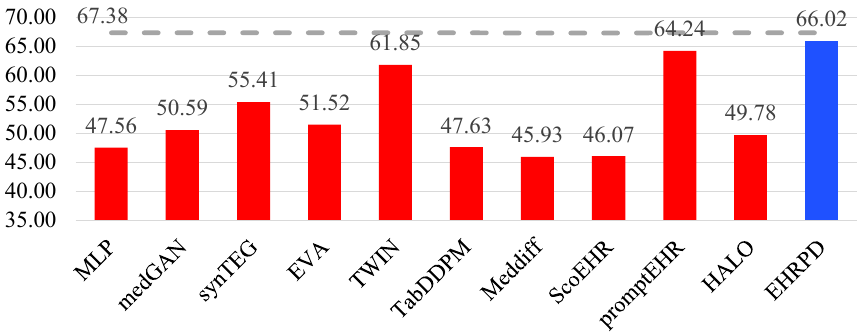}
    \vspace{-0.1in}
    \caption{Illustration of adverse event prediction with synthetic datasets.}
    \label{fig:aeplot}
    \vspace{-0.2in}
\end{figure}

Our findings are visually represented in Figure~\ref{fig:aeplot}, where the horizontal dashed line indicates the performance with the real training set. Red histograms show the performance of baseline models, while blue ones highlight that of our model. We can observe that our model is closest to the real dataset's performance, while PromptEHR achieves the second-best performance, likely due to the sequential generation nature of the model design that helps preserve the visit-to-visit consistency. However, HALO behaves worse than expected. This can be attributed to its design, which generates a single prediction vector of various modalities, diluting its effectiveness in tasks that require a focused prediction on a singular modality.

\begin{table}[t]
\resizebox{1\columnwidth}{!}{
\begin{tabular}{l|cc|cc|cc|cc}
\toprule
& \multicolumn{2}{c|}{Diagnosis} & \multicolumn{2}{c|}{Drug} & \multicolumn{2}{c|}{Lab Item} & \multicolumn{2}{c}{Procedure} \\ \hline
Metric & LPL          & MPL         & LPL           & MPL          & LPL          & MPL          & LPL          & MPL          \\ \hline
AS1    &22.60&26.99&32.84&36.69&23.68&20.81&34.21&33.22\\
AS2    &21.09&22.63&23.38&24.02&17.36&17.21&23.20&28.64\\
AS3    &22.96&22.71&28.30&27.72&19.36&18.46&18.95&22.52\\
AS4    &66.73&49.43&44.71&30.10&52.25&27.64&58.36&170.66\\ \hline
\ours   &\textbf{15.97}&\textbf{17.95}&\textbf{20.53}&\textbf{19.15}&\textbf{15.11}&\textbf{13.99}&\textbf{14.53}&\textbf{18.89}  \\
\bottomrule
\end{tabular}
}
\caption{Results of ablation study}
\label{table:ablation}
\vspace{-0.25in}
\end{table}

\subsection{Ablation Study}

In this section, we remove some components of our model to assess each component's effectiveness towards the whole model, with LPL and MPL as evaluation metrics. All ablation experiments are described as follows: 

\begin{itemize}
    \item \textbf{AS 1:} removes the time aware visit embedding in Section~\ref{sec:code-visit-embedding} and replaces with a linear embedding layer.
    \item \textbf{AS 2:} removes the time interval estimation (Eq.~\ref{eq:timepred}) and time prediction loss (Eq.~\ref{timeLoss}).
    \item \textbf{AS 3:} removes the demographic information embedding of catalyst representation in Section~\ref{sec:catalyst_representation}.
    \item \textbf{AS 4:} removes the self attention in Eq.~\eqref{selfatt}, i.e., exclude catalyst representation entirely.
\end{itemize}
The experiment result is shown in Table~\ref{table:ablation}. An analysis of the outcomes reveals that each component plays a significant role in enhancing the model's performance. Notably, the catalyst representation in \ours emerges as the most critical element, significantly influencing the model's performance.
\section{Conclusion}

In this paper, we present \ours, a diffusion-based EHR data generation model. By incorporating a time-aware visit embedding module and predicting the next visit with a novel predictive diffusion model, \ours is capable of capturing the complex temporal information of EHR data. Furthermore, \ours's ability to simultaneously estimate time intervals till the next visit sets it apart from existing methods, offering a significant improvement in the field of EHR data generation. To validate our claims, we conducted extensive experiments on publically available datasets, demonstrating \ours's superior performance from three comprehensive perspectives: utility, fidelity, and privacy.


\section*{Acknowledgements}
The authors thank the anonymous referees for their valuable comments and helpful suggestions. This work is partially supported by the National Science Foundation under Grant No. 2238275 and the National Institutes of Health under Grant No. R01AG077016.

\newpage
\bibliographystyle{ACM-Reference-Format}
\bibliography{bibfile.bib}


\begin{thebibliography}{38}


\ifx \showCODEN    \undefined \def \showCODEN     #1{\unskip}     \fi
\ifx \showDOI      \undefined \def \showDOI       #1{#1}\fi
\ifx \showISBNx    \undefined \def \showISBNx     #1{\unskip}     \fi
\ifx \showISBNxiii \undefined \def \showISBNxiii  #1{\unskip}     \fi
\ifx \showISSN     \undefined \def \showISSN      #1{\unskip}     \fi
\ifx \showLCCN     \undefined \def \showLCCN      #1{\unskip}     \fi
\ifx \shownote     \undefined \def \shownote      #1{#1}          \fi
\ifx \showarticletitle \undefined \def \showarticletitle #1{#1}   \fi
\ifx \showURL      \undefined \def \showURL       {\relax}        \fi
\providecommand\bibfield[2]{#2}
\providecommand\bibinfo[2]{#2}
\providecommand\natexlab[1]{#1}
\providecommand\showeprint[2][]{arXiv:#2}

\bibitem[Austin et~al\mbox{.}(2021)]%
        {austin2021structured}
\bibfield{author}{\bibinfo{person}{Jacob Austin}, \bibinfo{person}{Daniel~D Johnson}, \bibinfo{person}{Jonathan Ho}, \bibinfo{person}{Daniel Tarlow}, {and} \bibinfo{person}{Rianne Van Den~Berg}.} \bibinfo{year}{2021}\natexlab{}.
\newblock \showarticletitle{Structured denoising diffusion models in discrete state-spaces}.
\newblock \bibinfo{journal}{\emph{Advances in Neural Information Processing Systems}} (\bibinfo{year}{2021}), \bibinfo{pages}{17981--17993}.
\newblock


\bibitem[Awad et~al\mbox{.}(2015)]%
        {svr}
\bibfield{author}{\bibinfo{person}{Mariette Awad}, \bibinfo{person}{Rahul Khanna}, \bibinfo{person}{Mariette Awad}, {and} \bibinfo{person}{Rahul Khanna}.} \bibinfo{year}{2015}\natexlab{}.
\newblock \showarticletitle{Support vector regression}.
\newblock \bibinfo{journal}{\emph{Efficient learning machines: Theories, concepts, and applications for engineers and system designers}} (\bibinfo{year}{2015}), \bibinfo{pages}{67--80}.
\newblock


\bibitem[Baowaly et~al\mbox{.}(2019)]%
        {medGAN}
\bibfield{author}{\bibinfo{person}{Mrinal~Kanti Baowaly}, \bibinfo{person}{Chia-Ching Lin}, \bibinfo{person}{Chao-Lin Liu}, {and} \bibinfo{person}{Kuan-Ta Chen}.} \bibinfo{year}{2019}\natexlab{}.
\newblock \showarticletitle{Synthesizing electronic health records using improved generative adversarial networks}.
\newblock \bibinfo{journal}{\emph{JAMIA}} (\bibinfo{year}{2019}), \bibinfo{pages}{228--241}.
\newblock


\bibitem[Biswal et~al\mbox{.}(2021)]%
        {eva}
\bibfield{author}{\bibinfo{person}{Siddharth Biswal}, \bibinfo{person}{Soumya Ghosh}, \bibinfo{person}{Jon Duke}, \bibinfo{person}{Bradley Malin}, \bibinfo{person}{Walter Stewart}, \bibinfo{person}{Cao Xiao}, {and} \bibinfo{person}{Jimeng Sun}.} \bibinfo{year}{2021}\natexlab{}.
\newblock \showarticletitle{EVA: Generating longitudinal electronic health records using conditional variational autoencoders}. In \bibinfo{booktitle}{\emph{Machine Learning for Healthcare Conference}}. \bibinfo{pages}{260--282}.
\newblock


\bibitem[Box and Pierce(1970)]%
        {arima}
\bibfield{author}{\bibinfo{person}{George~EP Box} {and} \bibinfo{person}{David~A Pierce}.} \bibinfo{year}{1970}\natexlab{}.
\newblock \showarticletitle{Distribution of residual autocorrelations in autoregressive-integrated moving average time series models}.
\newblock \bibinfo{journal}{\emph{J. Amer. Statist. Assoc.}} (\bibinfo{year}{1970}), \bibinfo{pages}{1509--1526}.
\newblock


\bibitem[Che et~al\mbox{.}(2017)]%
        {ehrGAN}
\bibfield{author}{\bibinfo{person}{Zhengping Che}, \bibinfo{person}{Yu Cheng}, \bibinfo{person}{Shuangfei Zhai}, \bibinfo{person}{Zhaonan Sun}, {and} \bibinfo{person}{Yan Liu}.} \bibinfo{year}{2017}\natexlab{}.
\newblock \showarticletitle{Boosting deep learning risk prediction with generative adversarial networks for electronic health records}. In \bibinfo{booktitle}{\emph{IEEE ICDM}}. \bibinfo{pages}{787--792}.
\newblock


\bibitem[Choi et~al\mbox{.}(2016)]%
        {retain}
\bibfield{author}{\bibinfo{person}{Edward Choi}, \bibinfo{person}{Mohammad~Taha Bahadori}, \bibinfo{person}{Jimeng Sun}, \bibinfo{person}{Joshua Kulas}, \bibinfo{person}{Andy Schuetz}, {and} \bibinfo{person}{Walter Stewart}.} \bibinfo{year}{2016}\natexlab{}.
\newblock \showarticletitle{Retain: An interpretable predictive model for healthcare using reverse time attention mechanism}.
\newblock \bibinfo{journal}{\emph{Advances in neural information processing systems}} (\bibinfo{year}{2016}).
\newblock


\bibitem[Das et~al\mbox{.}(2023)]%
        {twin}
\bibfield{author}{\bibinfo{person}{Trisha Das}, \bibinfo{person}{Zifeng Wang}, {and} \bibinfo{person}{Jimeng Sun}.} \bibinfo{year}{2023}\natexlab{}.
\newblock \showarticletitle{Twin: Personalized clinical trial digital twin generation}. In \bibinfo{booktitle}{\emph{ACM SIGKDD Conference on Knowledge Discovery and Data Mining}}. \bibinfo{pages}{402--413}.
\newblock


\bibitem[Feng et~al\mbox{.}(2019)]%
        {dcmn}
\bibfield{author}{\bibinfo{person}{Yujuan Feng}, \bibinfo{person}{Zhenxing Xu}, \bibinfo{person}{Lin Gan}, \bibinfo{person}{Ning Chen}, \bibinfo{person}{Bin Yu}, \bibinfo{person}{Ting Chen}, {and} \bibinfo{person}{Fei Wang}.} \bibinfo{year}{2019}\natexlab{}.
\newblock \showarticletitle{Dcmn: Double core memory network for patient outcome prediction with multimodal data}. In \bibinfo{booktitle}{\emph{International Conference on Data Mining}}. \bibinfo{pages}{200--209}.
\newblock


\bibitem[Gong et~al\mbox{.}(2022)]%
        {gong2022diffuseq}
\bibfield{author}{\bibinfo{person}{Shansan Gong}, \bibinfo{person}{Mukai Li}, \bibinfo{person}{Jiangtao Feng}, \bibinfo{person}{Zhiyong Wu}, {and} \bibinfo{person}{Lingpeng Kong}.} \bibinfo{year}{2022}\natexlab{}.
\newblock \showarticletitle{DiffuSeq: Sequence to Sequence Text Generation with Diffusion Models}. In \bibinfo{booktitle}{\emph{International Conference on Learning Representations}}.
\newblock


\bibitem[He et~al\mbox{.}(2023)]%
        {he2023meddiff}
\bibfield{author}{\bibinfo{person}{Huan He}, \bibinfo{person}{Shifan Zhao}, \bibinfo{person}{Yuanzhe Xi}, {and} \bibinfo{person}{Joyce~C Ho}.} \bibinfo{year}{2023}\natexlab{}.
\newblock \showarticletitle{MedDiff: Generating electronic health records using accelerated denoising diffusion model}.
\newblock \bibinfo{journal}{\emph{arXiv preprint arXiv:2302.04355}} (\bibinfo{year}{2023}).
\newblock


\bibitem[Ho et~al\mbox{.}(2020)]%
        {ddpm}
\bibfield{author}{\bibinfo{person}{Jonathan Ho}, \bibinfo{person}{Ajay Jain}, {and} \bibinfo{person}{Pieter Abbeel}.} \bibinfo{year}{2020}\natexlab{}.
\newblock \showarticletitle{Denoising diffusion probabilistic models}.
\newblock \bibinfo{journal}{\emph{Advances in neural information processing systems}} (\bibinfo{year}{2020}), \bibinfo{pages}{6840--6851}.
\newblock


\bibitem[Hochreiter and Schmidhuber(1997)]%
        {lstm}
\bibfield{author}{\bibinfo{person}{Sepp Hochreiter} {and} \bibinfo{person}{J{\"u}rgen Schmidhuber}.} \bibinfo{year}{1997}\natexlab{}.
\newblock \showarticletitle{Long short-term memory}.
\newblock \bibinfo{journal}{\emph{Neural computation}} (\bibinfo{year}{1997}), \bibinfo{pages}{1735--1780}.
\newblock


\bibitem[Johnson et~al\mbox{.}(2016)]%
        {mimiciii}
\bibfield{author}{\bibinfo{person}{Alistair~EW Johnson}, \bibinfo{person}{Tom~J Pollard}, \bibinfo{person}{Lu Shen}, \bibinfo{person}{Li-wei~H Lehman}, \bibinfo{person}{Mengling Feng}, \bibinfo{person}{Mohammad Ghassemi}, \bibinfo{person}{Benjamin Moody}, \bibinfo{person}{Peter Szolovits}, \bibinfo{person}{Leo Anthony~Celi}, {and} \bibinfo{person}{Roger~G Mark}.} \bibinfo{year}{2016}\natexlab{}.
\newblock \showarticletitle{MIMIC-III, a freely accessible critical care database}.
\newblock \bibinfo{journal}{\emph{Scientific data}} (\bibinfo{year}{2016}), \bibinfo{pages}{1--9}.
\newblock


\bibitem[Kotelnikov et~al\mbox{.}(2023)]%
        {tabddpm}
\bibfield{author}{\bibinfo{person}{Akim Kotelnikov}, \bibinfo{person}{Dmitry Baranchuk}, \bibinfo{person}{Ivan Rubachev}, {and} \bibinfo{person}{Artem Babenko}.} \bibinfo{year}{2023}\natexlab{}.
\newblock \showarticletitle{Tabddpm: Modelling tabular data with diffusion models}. In \bibinfo{booktitle}{\emph{International Conference on Machine Learning}}. \bibinfo{pages}{17564--17579}.
\newblock


\bibitem[Li et~al\mbox{.}(2022b)]%
        {li2022srdiff}
\bibfield{author}{\bibinfo{person}{Haoying Li}, \bibinfo{person}{Yifan Yang}, \bibinfo{person}{Meng Chang}, \bibinfo{person}{Shiqi Chen}, \bibinfo{person}{Huajun Feng}, \bibinfo{person}{Zhihai Xu}, \bibinfo{person}{Qi Li}, {and} \bibinfo{person}{Yueting Chen}.} \bibinfo{year}{2022}\natexlab{b}.
\newblock \showarticletitle{Srdiff: Single image super-resolution with diffusion probabilistic models}.
\newblock \bibinfo{journal}{\emph{Neurocomputing}} (\bibinfo{year}{2022}), \bibinfo{pages}{47--59}.
\newblock


\bibitem[Li et~al\mbox{.}(2022a)]%
        {li2022diffusion}
\bibfield{author}{\bibinfo{person}{Xiang Li}, \bibinfo{person}{John Thickstun}, \bibinfo{person}{Ishaan Gulrajani}, \bibinfo{person}{Percy~S Liang}, {and} \bibinfo{person}{Tatsunori~B Hashimoto}.} \bibinfo{year}{2022}\natexlab{a}.
\newblock \showarticletitle{Diffusion-lm improves controllable text generation}.
\newblock \bibinfo{journal}{\emph{Advances in Neural Information Processing Systems}} (\bibinfo{year}{2022}), \bibinfo{pages}{4328--4343}.
\newblock


\bibitem[Luo et~al\mbox{.}(2020)]%
        {hitanet}
\bibfield{author}{\bibinfo{person}{Junyu Luo}, \bibinfo{person}{Muchao Ye}, \bibinfo{person}{Cao Xiao}, {and} \bibinfo{person}{Fenglong Ma}.} \bibinfo{year}{2020}\natexlab{}.
\newblock \showarticletitle{Hitanet: Hierarchical time-aware attention networks for risk prediction on electronic health records}. In \bibinfo{booktitle}{\emph{ACM SIGKDD International Conference on Knowledge Discovery and Data Mining}}. \bibinfo{pages}{647--656}.
\newblock


\bibitem[Ma et~al\mbox{.}(2017)]%
        {dipole}
\bibfield{author}{\bibinfo{person}{Fenglong Ma}, \bibinfo{person}{Radha Chitta}, \bibinfo{person}{Jing Zhou}, \bibinfo{person}{Quanzeng You}, \bibinfo{person}{Tong Sun}, {and} \bibinfo{person}{Jing Gao}.} \bibinfo{year}{2017}\natexlab{}.
\newblock \showarticletitle{Dipole: Diagnosis prediction in healthcare via attention-based bidirectional recurrent neural networks}. In \bibinfo{booktitle}{\emph{ACM SIGKDD International Conference on Knowledge Discovery and Data Mining}}. \bibinfo{pages}{1903--1911}.
\newblock


\bibitem[Ma et~al\mbox{.}(2020)]%
        {adacare}
\bibfield{author}{\bibinfo{person}{Liantao Ma}, \bibinfo{person}{Junyi Gao}, \bibinfo{person}{Yasha Wang}, \bibinfo{person}{Chaohe Zhang}, \bibinfo{person}{Jiangtao Wang}, \bibinfo{person}{Wenjie Ruan}, \bibinfo{person}{Wen Tang}, \bibinfo{person}{Xin Gao}, {and} \bibinfo{person}{Xinyu Ma}.} \bibinfo{year}{2020}\natexlab{}.
\newblock \showarticletitle{Adacare: Explainable clinical health status representation learning via scale-adaptive feature extraction and recalibration}. In \bibinfo{booktitle}{\emph{Association for the Advancement of Artificial Intelligence}}. \bibinfo{pages}{825--832}.
\newblock


\bibitem[Mei and Eisner(2017)]%
        {hawkeslstm}
\bibfield{author}{\bibinfo{person}{Hongyuan Mei} {and} \bibinfo{person}{Jason~M Eisner}.} \bibinfo{year}{2017}\natexlab{}.
\newblock \showarticletitle{The neural hawkes process: A neurally self-modulating multivariate point process}.
\newblock \bibinfo{journal}{\emph{Advances in neural information processing systems}} (\bibinfo{year}{2017}).
\newblock


\bibitem[Naseer et~al\mbox{.}(2023)]%
        {naseer2023scoehr}
\bibfield{author}{\bibinfo{person}{Ahmed~Ammar Naseer}, \bibinfo{person}{Benjamin Walker}, \bibinfo{person}{Christopher Landon}, \bibinfo{person}{Andrew Ambrosy}, \bibinfo{person}{Marat Fudim}, \bibinfo{person}{Nicholas Wysham}, \bibinfo{person}{Botros Toro}, \bibinfo{person}{Sumanth Swaminathan}, {and} \bibinfo{person}{Terry Lyons}.} \bibinfo{year}{2023}\natexlab{}.
\newblock \showarticletitle{ScoEHR: Generating Synthetic Electronic Health Records using Continuous-time Diffusion Models}. In \bibinfo{booktitle}{\emph{Machine Learning for Healthcare Conference}}. PMLR, \bibinfo{pages}{489--508}.
\newblock


\bibitem[Nichol et~al\mbox{.}(2022)]%
        {glide}
\bibfield{author}{\bibinfo{person}{Alexander~Quinn Nichol}, \bibinfo{person}{Prafulla Dhariwal}, \bibinfo{person}{Aditya Ramesh}, \bibinfo{person}{Pranav Shyam}, \bibinfo{person}{Pamela Mishkin}, \bibinfo{person}{Bob Mcgrew}, \bibinfo{person}{Ilya Sutskever}, {and} \bibinfo{person}{Mark Chen}.} \bibinfo{year}{2022}\natexlab{}.
\newblock \showarticletitle{GLIDE: Towards Photorealistic Image Generation and Editing with Text-Guided Diffusion Models}. In \bibinfo{booktitle}{\emph{International Conference on Machine Learning}}. \bibinfo{pages}{16784--16804}.
\newblock


\bibitem[Rasul et~al\mbox{.}(2021)]%
        {rasul2021autoregressive}
\bibfield{author}{\bibinfo{person}{Kashif Rasul}, \bibinfo{person}{Calvin Seward}, \bibinfo{person}{Ingmar Schuster}, {and} \bibinfo{person}{Roland Vollgraf}.} \bibinfo{year}{2021}\natexlab{}.
\newblock \showarticletitle{Autoregressive denoising diffusion models for multivariate probabilistic time series forecasting}. In \bibinfo{booktitle}{\emph{International Conference on Machine Learning}}. \bibinfo{pages}{8857--8868}.
\newblock


\bibitem[Rombach et~al\mbox{.}(2022)]%
        {rombach2022high}
\bibfield{author}{\bibinfo{person}{Robin Rombach}, \bibinfo{person}{Andreas Blattmann}, \bibinfo{person}{Dominik Lorenz}, \bibinfo{person}{Patrick Esser}, {and} \bibinfo{person}{Bj{\"o}rn Ommer}.} \bibinfo{year}{2022}\natexlab{}.
\newblock \showarticletitle{High-resolution image synthesis with latent diffusion models}. In \bibinfo{booktitle}{\emph{Conference on Computer Vision and Pattern Recognition}}. \bibinfo{pages}{10684--10695}.
\newblock


\bibitem[Saharia et~al\mbox{.}(2022)]%
        {saharia2022image}
\bibfield{author}{\bibinfo{person}{Chitwan Saharia}, \bibinfo{person}{Jonathan Ho}, \bibinfo{person}{William Chan}, \bibinfo{person}{Tim Salimans}, \bibinfo{person}{David~J Fleet}, {and} \bibinfo{person}{Mohammad Norouzi}.} \bibinfo{year}{2022}\natexlab{}.
\newblock \showarticletitle{Image super-resolution via iterative refinement}.
\newblock \bibinfo{journal}{\emph{IEEE Transactions on Pattern Analysis and Machine Intelligence}} (\bibinfo{year}{2022}), \bibinfo{pages}{4713--4726}.
\newblock


\bibitem[Schapire(2003)]%
        {gbrt}
\bibfield{author}{\bibinfo{person}{Robert~E Schapire}.} \bibinfo{year}{2003}\natexlab{}.
\newblock \showarticletitle{The boosting approach to machine learning: An overview}.
\newblock \bibinfo{journal}{\emph{Nonlinear estimation and classification}} (\bibinfo{year}{2003}), \bibinfo{pages}{149--171}.
\newblock


\bibitem[Tang et~al\mbox{.}(2020)]%
        {flstmandfcnn}
\bibfield{author}{\bibinfo{person}{Shengpu Tang}, \bibinfo{person}{Parmida Davarmanesh}, \bibinfo{person}{Yanmeng Song}, \bibinfo{person}{Danai Koutra}, \bibinfo{person}{Michael~W Sjoding}, {and} \bibinfo{person}{Jenna Wiens}.} \bibinfo{year}{2020}\natexlab{}.
\newblock \showarticletitle{Democratizing EHR analyses with FIDDLE: a flexible data-driven preprocessing pipeline for structured clinical data}.
\newblock \bibinfo{journal}{\emph{Journal of the American Medical Informatics Association}} (\bibinfo{year}{2020}), \bibinfo{pages}{1921--1934}.
\newblock


\bibitem[Tashiro et~al\mbox{.}(2021)]%
        {tashiro2021csdi}
\bibfield{author}{\bibinfo{person}{Yusuke Tashiro}, \bibinfo{person}{Jiaming Song}, \bibinfo{person}{Yang Song}, {and} \bibinfo{person}{Stefano Ermon}.} \bibinfo{year}{2021}\natexlab{}.
\newblock \showarticletitle{Csdi: Conditional score-based diffusion models for probabilistic time series imputation}.
\newblock \bibinfo{journal}{\emph{Advances in Neural Information Processing Systems}} (\bibinfo{year}{2021}), \bibinfo{pages}{24804--24816}.
\newblock


\bibitem[Theodorou et~al\mbox{.}(2023)]%
        {HALO}
\bibfield{author}{\bibinfo{person}{Brandon Theodorou}, \bibinfo{person}{Cao Xiao}, {and} \bibinfo{person}{Jimeng Sun}.} \bibinfo{year}{2023}\natexlab{}.
\newblock \showarticletitle{Synthesize high-dimensional longitudinal electronic health records via hierarchical autoregressive language model}.
\newblock \bibinfo{journal}{\emph{Nature communications}} (\bibinfo{year}{2023}), \bibinfo{pages}{5305}.
\newblock


\bibitem[Wang et~al\mbox{.}(2023)]%
        {medhmp}
\bibfield{author}{\bibinfo{person}{Xiaochen Wang}, \bibinfo{person}{Junyu Luo}, \bibinfo{person}{Jiaqi Wang}, \bibinfo{person}{Ziyi Yin}, \bibinfo{person}{Suhan Cui}, \bibinfo{person}{Yuan Zhong}, \bibinfo{person}{Yaqing Wang}, {and} \bibinfo{person}{Fenglong Ma}.} \bibinfo{year}{2023}\natexlab{}.
\newblock \showarticletitle{Hierarchical Pretraining on Multimodal Electronic Health Records}. In \bibinfo{booktitle}{\emph{Empirical Methods in Natural Language Processing}}.
\newblock


\bibitem[Wang and Sun(2022)]%
        {promptehr}
\bibfield{author}{\bibinfo{person}{Zifeng Wang} {and} \bibinfo{person}{Jimeng Sun}.} \bibinfo{year}{2022}\natexlab{}.
\newblock \showarticletitle{PromptEHR: Conditional Electronic Healthcare Records Generation with Prompt Learning}. In \bibinfo{booktitle}{\emph{Conference on Empirical Methods in Natural Language Processing}}.
\newblock


\bibitem[Welch et~al\mbox{.}(1995)]%
        {kalman}
\bibfield{author}{\bibinfo{person}{Greg Welch}, \bibinfo{person}{Gary Bishop}, {et~al\mbox{.}}} \bibinfo{year}{1995}\natexlab{}.
\newblock \showarticletitle{An introduction to the Kalman filter}.
\newblock  (\bibinfo{year}{1995}).
\newblock


\bibitem[Xiao et~al\mbox{.}(2018)]%
        {xiao2018opportunities}
\bibfield{author}{\bibinfo{person}{Cao Xiao}, \bibinfo{person}{Edward Choi}, {and} \bibinfo{person}{Jimeng Sun}.} \bibinfo{year}{2018}\natexlab{}.
\newblock \showarticletitle{Opportunities and challenges in developing deep learning models using electronic health records data: a systematic review}.
\newblock \bibinfo{journal}{\emph{Journal of the American Medical Informatics Association}} (\bibinfo{year}{2018}), \bibinfo{pages}{1419--1428}.
\newblock


\bibitem[Xu et~al\mbox{.}(2018)]%
        {raim}
\bibfield{author}{\bibinfo{person}{Yanbo Xu}, \bibinfo{person}{Siddharth Biswal}, \bibinfo{person}{Shriprasad~R Deshpande}, \bibinfo{person}{Kevin~O Maher}, {and} \bibinfo{person}{Jimeng Sun}.} \bibinfo{year}{2018}\natexlab{}.
\newblock \showarticletitle{Raim: Recurrent attentive and intensive model of multimodal patient monitoring data}. In \bibinfo{booktitle}{\emph{ACM SIGKDD international conference on Knowledge Discovery and Data Mining}}. \bibinfo{pages}{2565--2573}.
\newblock


\bibitem[Zhang et~al\mbox{.}(2021)]%
        {synteg}
\bibfield{author}{\bibinfo{person}{Ziqi Zhang}, \bibinfo{person}{Chao Yan}, \bibinfo{person}{Thomas~A Lasko}, \bibinfo{person}{Jimeng Sun}, {and} \bibinfo{person}{Bradley~A Malin}.} \bibinfo{year}{2021}\natexlab{}.
\newblock \showarticletitle{SynTEG: a framework for temporal structured electronic health data simulation}.
\newblock \bibinfo{journal}{\emph{Journal of the American Medical Informatics Association}} (\bibinfo{year}{2021}), \bibinfo{pages}{596--604}.
\newblock


\bibitem[Zhong et~al\mbox{.}(2024)]%
        {meddiffusion}
\bibfield{author}{\bibinfo{person}{Yuan Zhong}, \bibinfo{person}{Suhan Cui}, \bibinfo{person}{Jiaqi Wang}, \bibinfo{person}{Xiaochen Wang}, \bibinfo{person}{Ziyi Yin}, \bibinfo{person}{Yaqing Wang}, \bibinfo{person}{Houping Xiao}, \bibinfo{person}{Mengdi Huai}, \bibinfo{person}{Ting Wang}, {and} \bibinfo{person}{Fenglong Ma}.} \bibinfo{year}{2024}\natexlab{}.
\newblock \showarticletitle{MedDiffusion: Boosting Health Risk Prediction via Diffusion-based Data Augmentation}. In \bibinfo{booktitle}{\emph{SIAM International Conference on Data Mining}}.
\newblock


\bibitem[Zhou et~al\mbox{.}(2021)]%
        {DBLP:conf/kdd/ZhouXWNKAH21}
\bibfield{author}{\bibinfo{person}{Yao Zhou}, \bibinfo{person}{Jianpeng Xu}, \bibinfo{person}{Jun Wu}, \bibinfo{person}{Zeinab~Taghavi Nasrabadi}, \bibinfo{person}{Evren K{\"{o}}rpeoglu}, \bibinfo{person}{Kannan Achan}, {and} \bibinfo{person}{Jingrui He}.} \bibinfo{year}{2021}\natexlab{}.
\newblock \showarticletitle{{PURE:} Positive-Unlabeled Recommendation with Generative Adversarial Network}. In \bibinfo{booktitle}{\emph{{KDD} '21: The 27th {ACM} {SIGKDD} Conference on Knowledge Discovery and Data Mining, Virtual Event, Singapore, August 14-18, 2021}}. \bibinfo{publisher}{{ACM}}, \bibinfo{pages}{2409--2419}.
\newblock


\end{thebibliography}

\section{Appendix}

\subsection{Details of \dmodel}

In this section, we provide formula derivations for the theoretical foundation of \dmodel.

\subsubsection{Forward Noise Addition Process}\label{apd:forward}

In the forward diffusion process of \dmodel, we gradually add noise to $\mathbf{v}_i$ according to the noise schedule $\mathbf{\beta}_s$:
\begin{equation}
    \begin{gathered}
    q(\mathbf{v}_i^{1:S}|\mathbf{v}_i^0) = \prod_{s=1}^S q(\mathbf{v}_i^s|\mathbf{v}_i^{s-1}),\\
    q(\mathbf{v}_i^s|\mathbf{v}_i^{s-1}) = \mathcal{N}(\mathbf{v}_i^s;\sqrt{1-\beta_s} \mathbf{v}_i^{s-1},\beta_s\mathbf{I}).
\end{gathered}
\end{equation}
Let $\alpha_s = 1-\beta_s$ and $\bar{\alpha}_s = \prod_{j=1}^s\alpha_j$, we can re-parameterize the Gaussian step above with its mean and variance as:
\begin{equation}
\begin{split}
\mathbf{v}_i^s =& \sqrt{\alpha_s} \mathbf{v}_i^{s-1} + \sqrt{1-\alpha_s}\epsilon_{s-1} \\
=& \sqrt{\alpha_s \alpha_{s-1}} \mathbf{v}_i^{s-2} + \sqrt{1-\alpha_s \alpha_{s-1}}\bar{\epsilon}_{s-2}\\
=& \cdots \\
=& \sqrt{\bar{\alpha}_s} \mathbf{v}_i^0 + \sqrt{1-\bar{\alpha}_s}\epsilon,
\end{split}
\end{equation}
where $\epsilon$ is the merged Gaussian noise term from $[\epsilon_1,\cdots,\epsilon_S]$ by the property of normal distribution.

\subsubsection{Predictive Mapping Process}

By Eq.\eqref{eq:predictive_mapping}, we construct a relationship between $\mathbf{v}_{i+1}^s$, $\mathbf{v}^s_i$, and $\mathbf{\Phi}_i$.

\subsubsection{Backward Denoising Diffusion Process}\label{apd:backward}

Then in the backward diffusion process, we start with $\mathbf{v}^s_{i+1}$. We utilize Bayes Theorem to rewrite the backward diffusion step into a mixture of forward Gaussian steps as:
\begin{equation}
\begin{split}
q(\mathbf{v}_{i+1}^{s-1}|\mathbf{v}_{i+1}^s,\mathbf{v}_{i+1}^0)
=q(\mathbf{v}_{i+1}^s|\mathbf{v}_{i+1}^{s-1},\mathbf{v}_{i+1}^0)\frac{q(\mathbf{v}_{i+1}^{s-1}|\mathbf{v}_{i+1}^0)}{q(\mathbf{v}_{i+1}^s|\mathbf{v}_{i+1}^0)}.
\end{split}
\end{equation}
Then by the density function of normal distribution, the above equation is proportional to:
\begin{equation}
\begin{split}
\propto& \exp \{ -\frac{1}{2} (
\frac{(\mathbf{v}_{i+1}^s-\sqrt{\alpha_s}\mathbf{v}_{i+1}^{s-1})^2}{\beta_s}
+ \frac{(\mathbf{v}_{i+1}^{s-1}-\sqrt{\bar{\alpha}_{s-1}}\mathbf{v}_{i+1}^0)^2}{1-\bar{\alpha}_{s-1}}\\
&- \frac{(\mathbf{v}_{i+1}^s-\sqrt{\bar{\alpha}_s}\mathbf{v}_{i+1}^0)^2}{1-\bar{\alpha_s}}
) \}\\
=& \exp\{ -\frac{1}{2} (
\frac{(\mathbf{v}_{i+1}^s)^2-2\sqrt{\alpha_s}\mathbf{v}_{i+1}^s\mathbf{v}_{i+1}^{s-1}+\alpha_s(\mathbf{v}_{i+1}^{s-1})^2}{\beta_s}\\
&+ \frac{(\mathbf{v}_{i+1}^{s-1})^2-2\sqrt{\bar{\alpha}_{t-1}}\mathbf{v}_{i+1}^0\mathbf{v}_{i+1}^{s-1}+\bar{\alpha}_{s-1}(\mathbf{v}_{i+1}^0)^2}{1-\bar{\alpha}_{s-1}}\\
&-\frac{(\mathbf{v}_{i+1}^s-\sqrt{\bar{\alpha}_s}\mathbf{v}^0_{i+1})^2}{1-\bar{\alpha}_s})\}\\
=&\exp\{-\frac{1}{2}((\frac{\alpha_s}{\beta_s}+\frac{1}{1-\bar{\alpha}_{s-1}})(\mathbf{v}^{s-1}_{i+1})^2\\
&- (\frac{2\sqrt{\alpha_s}}{\beta_s}\mathbf{v}_{i+1}^s+\frac{2\sqrt{\bar{\alpha}_{s-1}}}{1-\bar{\alpha}_{s-1}}\mathbf{v}_{i+1}^0)\mathbf{v}_{i+1}^{s-1} + \text{Constant}
) \}
\end{split}
\end{equation}
Then by inspection, we can derive the mean and variance of the above density function as:
\begin{equation}
\begin{split}
&\hat{\beta}_s 
= 1/(\frac{\alpha_s}{\beta_s}+\frac{1}{1-\bar{\alpha}_{s-1}})
=\frac{1-\bar{\alpha}_{s-1}}{1-\bar{\alpha}_s}\beta_s,\\
&\hat{\boldsymbol{\mu}}_s(\mathbf{v}_{i+1}^s,\mathbf{v}_{i+1}^0)\\
=&(\frac{\sqrt{\alpha_s}}{\beta_s}\mathbf{v}_{i+1}^s+\frac{\sqrt{\bar{\alpha}_{s-1}}}{1-\bar{\alpha}_{s-1}}\mathbf{v}_{i+1}^0)/(\frac{\alpha_s}{\beta_s}+\frac{1}{1-\bar{\alpha}_{s-1}})    \\
=&(\frac{\sqrt{\alpha_s}}{\beta_s}\mathbf{v}_{i+1}^s+\frac{\sqrt{\bar{\alpha}_{s-1}}}{1-\bar{\alpha}_{s-1}}\mathbf{v}_{i+1}^0)\frac{1-\bar{\alpha}_{s-1}}{1-\bar{\alpha}_s}\beta_s \\
=&\frac{\sqrt{\alpha_s}(1-\bar{\alpha}_{s-1})}{1-\bar{\mathbf{\alpha}_s}}  \mathbf{v}_{i+1}^s+ \frac{\sqrt{\bar{\alpha}_{s-1}}\beta_s}{1-\bar{\alpha}_s}\mathbf{v}_{i+1}^0.
\end{split}
\end{equation}
Substituting $\mathbf{v}_{i+1}^0$ with Eq.\eqref{eq:forwardRepa} by $\mathbf{v}_{i+1}^s$, we have:

\begin{equation}
\begin{split}
&\hat{\boldsymbol{\mu}}_s(\mathbf{v}_{i+1}^s,\mathbf{v}_{i+1}^0) \\
=&\frac{\sqrt{\alpha_s}(1-\bar{\alpha}_{s-1})}{1-\bar{\mathbf{\alpha}_s}}  \mathbf{v}_{i+1}^s+ \frac{\sqrt{\bar{\alpha}_{s-1}}\beta_s}{1-\bar{\alpha}_s} \frac{1}{\sqrt{\bar{\alpha}_s}}( \mathbf{v}_{i+1}^s - \sqrt{1-\bar{\alpha}_s}\epsilon_s) \\
=&\frac{1}{\sqrt{\alpha}_t}(\mathbf{v}_{i+1}^s - \frac{1-\alpha_s}{\sqrt{1-\bar{\alpha_t}}}\epsilon_s)
\end{split}
\end{equation}
And finally we have the closed-form solution that describes the cross-visit relation with Eq.~\eqref{eq:predictive_mapping} as follows:
\begin{equation}   \hat{\boldsymbol{\mu}}_s(\mathbf{v}_{i+1}^s,\mathbf{v}_{i+1}^0) = \frac{1}{\sqrt{\alpha}_s}(f(\mathbf{v}_i^s, \boldsymbol{\Phi}_i) - \frac{1-\alpha_s}{\sqrt{1-\bar{\alpha}_s}}\epsilon_s).
\label{apd:mapped-meanAndVarofModel1}
\end{equation}


\subsection{Details of \unet}\label{apd:unet}


In this section, we provide the detailed structure of our \unet, as in Figure~\ref{fig:unet}.

\subsubsection{Downsampling Path.}

Denoting the BatchNorm layer as BN, the downsampling path of the \unet utilizes Resnet Block (ResB) to refine features and downsamples with a 1-D convolutional layer as follows:
\begin{equation}
    \begin{gathered}
        \text{ResB}(\mathbf{v}_{i,l}^s) = \text{ReLU}(\text{BN}(\text{Conv}(\mathbf{v}_{i,l}^s))) + \mathbf{v}_{i,l}^s,\\
        \mathbf{v}_{i,l+1}^s = \text{Conv1d}(\text{ResB}(\mathbf{v}_{i,l}^s)).
    \end{gathered}
\end{equation}

\subsubsection{Self-attention}

In the skip connection, we utilize a self-attention to fuse $l$-th layer $\boldsymbol{\Phi}_{i,l}$ and $\mathbf{v}_{i,l}^{s}$:
\begin{equation}
\begin{gathered}
\bar{\mathbf{v}}_{i+1,l}^{s} = \text{Softmax}\left(\frac{\mathbf{W}^Q_l(\mathbf{v}_{i,l}^{s})\cdot\mathbf{W}^K_l(\boldsymbol{\Phi}_{i,l})}{\sqrt{d}}\right)\cdot\mathbf{W}^V_l(\boldsymbol{\Phi}_{i,l}),\\
\dot{\mathbf{v}}_{i+1,l}^{s} = \text{MaxPooling}(\text{LayerNorm}(\mathbf{v}_{i,l}^{s}) + \bar{\mathbf{v}}_{i+1,l}^{s}),
\end{gathered}
\end{equation}
where $\mathbf{W}^Q_l, \mathbf{W}^K_l, \mathbf{W}^V_l \in \mathbb{R}^{d_l*d_l}$.

\subsubsection{Upsampling Path}

With DeConv denoting the deconvolution layer, our upsampling path first upsamples the lower level feature $\mathbf{v}_{i+1,l+1}^s$ to $\ddot{\mathbf{v}}_{i+1,l}^s$:
\begin{equation}
    \ddot{\mathbf{v}}_{i+1,l}^s = \text{ReLU}(\text{BN}(\text{DeConv1d}(\mathbf{v}_{i+1,l+1}^s))).
\end{equation}
Then the feature from the skip connection is fused with the upsampled feature with a 1-D convolution layer as:
\begin{equation}
    \tilde{\mathbf{v}}_{i+1,l}^s= \text{Conv1d}[\dot{\mathbf{v}}_{i+1,l}^s;\ddot{\mathbf{v}}_{i+1,l}^s].
\end{equation}
Lastly, we utilize a Resnet block to refine the learned feature:
\begin{equation}
    \hat{\mathbf{v}}_{i+1,l}^s = \text{ResB}(\tilde{\mathbf{v}}_{i+1,l}^s).
\end{equation}

\begin{table*}[h]
\resizebox{0.9\textwidth}{!}
{
\begin{tabular}{c|l|cccccccccccc}
\toprule
\multirow{2}{*}{\textbf{Task}}& 
\textbf{Model }         & \multicolumn{3}{c|}{F-LSTM}         & \multicolumn{3}{c|}{F-CNN}              & \multicolumn{3}{c|}{RAIM}               & \multicolumn{3}{c}{DCMN} \\ \cline{2-14}
&\textbf{ Metric}  & AUPR & F1 & \multicolumn{1}{c|}{Kappa} & AUPR & F1 & \multicolumn{1}{c|}{Kappa} & AUPR & F1 & \multicolumn{1}{c|}{Kappa} & AUPR     & F1     & Kappa     \\ \hline
\multirow{10}{*}{\rotatebox{90}{ARF}}
&Orginal       &0.9582&0.8969& \multicolumn{1}{c|}{0.7826}      &0.9550&0.8794& \multicolumn{1}{c|}{0.7590}      &0.9465&0.8698& \multicolumn{1}{c|}{0.7307}      &0.9471&0.8795&0.7439  \\
&MLP      &0.9577&0.8932& \multicolumn{1}{c|}{0.7810}      &0.9587&0.8886& \multicolumn{1}{c|}{0.7635}      &0.9494&0.8713& \multicolumn{1}{c|}{0.7419}      &0.9438&0.8756&0.7486  \\
&medGAN   &0.9518&0.8871& \multicolumn{1}{c|}{0.7610}      &0.9535&0.8873& \multicolumn{1}{c|}{0.7673}      &0.9538&0.8715& \multicolumn{1}{c|}{0.7408}      &0.9523&0.8754&0.7449  \\
&synTEG        &0.9590&0.8929& \multicolumn{1}{c|}{0.7722}      &0.9535&0.8812& \multicolumn{1}{c|}{0.7701}      &0.9445&0.8636& \multicolumn{1}{c|}{0.7273}      &0.9536&0.8871&0.7610  \\
&EVA           &0.9600&0.8980& \multicolumn{1}{c|}{0.7820}      &0.9526&0.8870& \multicolumn{1}{c|}{0.7623}      &0.9478&0.8761& \multicolumn{1}{c|}{0.7358}      &0.9530&0.8856&0.7663  \\
&TWIN          &0.9617&0.8997& \multicolumn{1}{c|}{0.7946}      &0.9537&0.8903& \multicolumn{1}{c|}{0.7728}      &0.9575&0.8820& \multicolumn{1}{c|}{0.7584}      &0.9541&0.8844&0.7629  \\ 
&TabDDPM  &0.9574&0.8952& \multicolumn{1}{c|}{0.7792}      &0.9567&0.8885& \multicolumn{1}{c|}{0.7720}      &0.9414&0.8706& \multicolumn{1}{c|}{0.7371}      &0.9518&0.8820&0.7559  \\
&Meddiff&0.9542&0.8965&\multicolumn{1}{c|}{0.7840}&0.9542&0.8837&\multicolumn{1}{c|}{0.7572}&0.9435&0.8700&\multicolumn{1}{c|}{0.7222}&0.9511&0.8824&0.7503\\
&ScoEHR&0.9559&0.9013&\multicolumn{1}{c|}{0.7933}&0.9487&0.8719&\multicolumn{1}{c|}{0.7114}&0.9487&0.8701&\multicolumn{1}{c|}{0.7260}&0.9524&0.8802&0.7521\\
&PromptEHR     &0.9530&0.8940& \multicolumn{1}{c|}{0.7706}      &0.9540&0.8797& \multicolumn{1}{c|}{0.7724}      &0.9562&0.8840& \multicolumn{1}{c|}{0.7443}      &0.9560&0.8894&0.7731  \\
&HALO     &\textbf{0.9645}&0.9020& \multicolumn{1}{c|}{0.7944}      &0.9546&0.8884& \multicolumn{1}{c|}{0.7657}      &0.9556&0.8776& \multicolumn{1}{c|}{0.7630}      &0.9542&0.8868&0.7657  \\ \cline{2-14}
&\ours         &0.9628&\textbf{0.9031}& \multicolumn{1}{l|}{\textbf{0.7975}}      &\textbf{0.9616}&\textbf{0.8918}& \multicolumn{1}{l|}{\textbf{0.7737}}      &\textbf{0.9582}&\textbf{0.8862}& \multicolumn{1}{l|}{\textbf{0.7637}}      &\textbf{0.9562}&\textbf{0.8931}&\textbf{0.7785}  \\ \hline

 \multirow{10}{*}{\rotatebox{90}{Shock}}
 &
Orginal       &0.8147&0.7092& \multicolumn{1}{l|}{0.5354}      &0.8110&0.7057& \multicolumn{1}{l|}{0.4669}      &0.8104&0.7455& \multicolumn{1}{l|}{0.5944}      &0.8012&0.7256&0.5726  \\
&MLP      &0.8274&0.7630& \multicolumn{1}{l|}{0.6298}      &0.8189&0.7500& \multicolumn{1}{l|}{0.5731}      &0.8101&0.7353& \multicolumn{1}{l|}{0.5868}      &0.8079&0.7345&0.5849  \\
&medGAN   &0.8379&0.7695& \multicolumn{1}{l|}{0.6322}      &0.8224&0.7579& \multicolumn{1}{l|}{0.5878}      &0.8010&0.7494& \multicolumn{1}{l|}{0.6005}      &0.8106&0.7399&0.5920  \\
&synTEG        &0.8391&0.7561& \multicolumn{1}{l|}{0.6155}      &0.8227&0.7513& \multicolumn{1}{l|}{0.6107}      &0.8262&0.7473& \multicolumn{1}{l|}{0.6053}      &0.8105&0.7449&0.5933  \\
&EVA           &0.8437&0.7638& \multicolumn{1}{l|}{0.6325}      &0.8324&0.7517& \multicolumn{1}{l|}{0.5934}      &0.8287&0.7406& \multicolumn{1}{l|}{0.6044}      &0.8192&0.7441&0.5908   \\
&TWIN          &0.8472&0.7639& \multicolumn{1}{l|}{0.6263}      &0.8382&0.7616& \multicolumn{1}{l|}{0.6282}      &0.8208&0.7374& \multicolumn{1}{l|}{0.5939}      &0.8280&0.7534&0.5968  \\ 
&TabDDPM  &0.8421&0.7684& \multicolumn{1}{l|}{0.6327}      &0.8247&0.7508& \multicolumn{1}{l|}{0.5882}      &0.8199&0.7477& \multicolumn{1}{l|}{0.6019}      &0.8085&0.7380&0.5824  \\
&Meddiff&0.8437&0.7651&\multicolumn{1}{c|}{0.6283}&0.8371&0.7651&\multicolumn{1}{c|}{0.6134}&0.8138&0.7456&\multicolumn{1}{c|}{0.5943
}&	0.8106&0.7423&0.5895\\
&ScoEHR&0.8477&0.7652&\multicolumn{1}{c|}{0.6400}&0.8316&0.7638&\multicolumn{1}{c|}{0.6148}&0.8153&0.7340&\multicolumn{1}{c|}{0.5826}&0.8117&0.7434&0.5872\\
&PromptEHR&0.8427&0.7661& \multicolumn{1}{l|}{0.6268}      &0.8334&0.7672& \multicolumn{1}{l|}{0.6179}      &0.8282&0.7437& \multicolumn{1}{l|}{0.5938}      &0.8152&0.7518&0.6060\\
&HALO     &\textbf{0.8563}&0.7709& \multicolumn{1}{l|}{\textbf{0.6419}}      &0.8253&0.7568& \multicolumn{1}{l|}{0.5923}      &0.8268&0.7538& \multicolumn{1}{l|}{0.6084}      &0.8212&0.7435&0.5926  \\ \cline{2-14}
&\ours          &0.8507&\textbf{0.7722}& \multicolumn{1}{l|}{0.6353}      &\textbf{0.8421}&\textbf{0.7704}& \multicolumn{1}{l|}{\textbf{0.6369}}      &\textbf{0.8361}&\textbf{0.7791}& \multicolumn{1}{l|}{\textbf{0.6189}}      &\textbf{0.8376}&\textbf{0.7704}&\textbf{0.6118} \\ 

\bottomrule
\end{tabular}
}
\caption{Result evaluation via other two risk prediction tasks on multimodal EHR data.}
\label{table:multimodal_apd}
\vspace{-0.2in}
\end{table*}

\begin{table*}[]
\centering
\resizebox{0.9\textwidth}{!}{
\begin{tabular}{c|c|c|c|c|c|c|c|c|c|c|c|c|c}
\toprule

\textbf{Dataset} & \textbf{Modality} & \textbf{Metric} & MLP    & medGAN & synTEG & EVA   & TWIN  & TabDDPM & Meddiff & ScoEHR & PromptEHR & HALO   & \ours \\ \hline
\multirow{8}{*}{\rotatebox{90}{MIMIC-III}}  & \multirow{2}{*}{Diagnosis}   & LPL             & 325.55 & 242.30 & 36.87  & 29.62 & 26.28 & 108.79  & 664.54  & 685.17 & 126.23    & 149.66 & \textbf{15.97}       \\
                 &                   & MPL             & 352.54 & 257.48 & 45.61  & 31.63 & 27.68 & 114.18  & 670.91  & 691.55 & 128.05    & 192.13 & \textbf{17.95}       \\\cline{2-14} 
                 & \multirow{2}{*}{Drug}              & LPL             & 553.63 & 403.02 & 83.38  & 43.79 & 40.94 & 179.22  & 936.28  & 934.87 & 167.48    & 166.11 & \textbf{20.53}       \\
                 &                   & MPL             & 551.67 & 405.75 & 82.66  & 44.02 & 40.86 & 178.70  & 936.14  & 950.27 & 136.04    & 202.01 & \textbf{19.15}       \\\cline{2-14} 
                 & \multirow{2}{*}{Lab Item}          & LPL             & 168.25 & 77.10  & 26.80  & 20.05 & 17.47 & 54.69   & 413.41  & 432.11 & 107.22    & 322.51 & \textbf{15.11}       \\
                 &                   & MPL             & 166.61 & 87.12  & 30.34  & 19.97 & 17.41 & 54.44   & 412.33  & 431.09 & 98.52     & 303.09 & \textbf{13.99}       \\\cline{2-14} 
                 & \multirow{2}{*}{Procedure}         & LPL             & 290.38 & 234.81 & 49.33  & 27.39 & 21.26 & 98.03   & 471.81  & 486.81 & 51.18     & 22.68  & \textbf{14.53}       \\
                 &                   & MPL             & 286.53 & 245.28 & 44.00  & 30.49 & 24.26 & 102.72  & 479.96  & 499.89 & 31.13     & 39.04  & \textbf{18.89}      \\

\bottomrule
\end{tabular}
}
\caption{EHR data generation evaluation of different approaches on eICU dataset.}
\label{table:eICU}
\vspace{-0.2in}
\end{table*}

\subsection{Baseline EHR Generation Models}\label{generationBaseline}

\begin{itemize}
    \item \textbf{MLP}~\cite{lstm} integrates an LSTM with an MLP to learn relationships between patient visits.
    \item \textbf{medGAN}~\cite{medGAN} uses a GAN to generate synthetic patient data, enhanced with an LSTM for temporal dynamics.
    \item \textbf{synTEG}~\cite{synteg} employs a Transformer for learning relationships in patient visit sequences and a Wasserstein GAN for generating EHR data sequences.
    \item \textbf{EVA}~\cite{eva} utilizes a VAE to encode health records into latent vectors and generate synthetic records from the learned distribution.
    \item \textbf{TWIN}~\cite{twin} combines a VAE for capturing data distribution with decoders for predicting current and next visit codes, focusing on cross-modality fusion and temporal dynamics.
    \item \textbf{TabDDPM}~\cite{tabddpm} generates tabular healthcare data, incorporating an LSTM for temporal learning.
    \item \textbf{Meddiff}~\cite{he2023meddiff} uses an accelerated DDPM to generate realistic synthetic EHR data, capturing temporal dependencies.
    \item \textbf{ScoEHR}~\cite{naseer2023scoehr} utilizes continuous-time diffusion models to generate synthetic EHR data with temporal dynamics.
    \item \textbf{PromptEHR}~\cite{promptehr} uses a pre-trained BART to generate diverse longitudinal EHR data.
    \item \textbf{HALO}~\cite{HALO} uses transformer architecture to learn different modalities of EHR codes jointly.

\end{itemize}

\subsection{Multimodal Risk Prediction Models}\label{apd:multimodalPrediction}

Multimodal risk prediction models used in Section~\ref{multiRiskPred}:

\begin{itemize}
    \item \textbf{F-LSTM}~\cite{flstmandfcnn} combines static demographic features with time-series features as input for an LSTM module.
    \item \textbf{F-CNN}~\cite{flstmandfcnn} is similar to F-LSTM but with a CNN.
    \item \textbf{RAIM}~\cite{raim} integrates attention mechanism with modality fusion and uses an LSTM for visit-wise relationship learning.
    \item \textbf{DCMN}~\cite{dcmn} utilizes separate recursive learning modules for each modality with an attention mechanism.
\end{itemize}

\subsection{Backbone Unimodal Risk Prediction Models}\label{apd:unimodalPrediction}

Unimodal risk prediction models used in Section~\ref{uniRiskPred}:

\begin{itemize}
    \item \textbf{AdaCare}~\cite{adacare} uses a CNN for feature extraction and GRU blocks for prediction. 
    \item \textbf{Dipole}~\cite{dipole} combines a bidirectional GRU with an attention mechanism to analyze patient visit sequences. 
    \item \textbf{HiTANet}~\cite{hitanet} adopts a time-aware attention mechanism to capture evolving disease patterns. 
    \item \textbf{LSTM}~\cite{lstm} learns the hidden state of each visit and performs risk prediction with an MLP.
    \item \textbf{Retain}~\cite{retain} employs a reverse time attention mechanism to prioritize recent medical events.
\end{itemize}

\subsection{Backbone Time Interval Prediction Models}\label{apd:timePrediction}

Time interval prediction methods in Section~\ref{timePred}:

\begin{itemize}
    \item \textbf{ARIMA}~\cite{arima} forecasts future values using past values and errors in a rolling window fashion.
    \item \textbf{KF}~\cite{kalman} estimates system states in linear dynamic systems.
    \item \textbf{SVR}~\cite{svr} predicts continuous values by fitting a regression line within an error margin.
    \item \textbf{GBRT}~\cite{gbrt} combines multiple decision trees to improve prediction accuracy through boosting.
    \item \textbf{LSTM}~\cite{lstm} outputs a single value for time prediction.
\end{itemize}

\subsection{More Result on eICU Dataset and Multimodal Risk Prediction Task}\label{apd:multiRiskPred}
Additional results on the eICU dataset and multimodal risk prediction task (Acute Respiratory Failure(ARF) and Shock) are shown in Table~\ref{table:eICU} and Table~\ref{table:multimodal_apd}.

\end{document}